\documentclass[a4paper,fleqn]{article}

\usepackage{arxiv}

\usepackage{hyperref}
\usepackage{natbib}
\usepackage{microtype}
\usepackage{graphicx}
\usepackage{subfigure}
\usepackage{booktabs} % for professional tables

\usepackage{url}
\usepackage{graphicx}
\usepackage{amsmath}
\usepackage{amssymb}
\usepackage{amsthm}
\usepackage{tikz}
\usepackage{mathrsfs}
\usepackage{bm}
\usepackage{verbatim}
\usepackage{setspace}
\usepackage{xcolor}
\usepackage{enumitem}
\usepackage[super]{nth}
\usepackage[utf8]{inputenc}
\usepackage{multirow}
\usepackage{hhline}
\usepackage{colortbl}
\usepackage{diagbox}
\usepackage{makecell}
\usepackage{tabularx}
\title{ENTMOOT: A Framework for Optimization over Ensemble Tree Models}
\usetikzlibrary{arrows}
\usetikzlibrary{arrows.meta}
\usetikzlibrary{calc}
\usetikzlibrary{decorations}
\usetikzlibrary{decorations.markings}
\usetikzlibrary{external}
\usetikzlibrary{patterns}
\usetikzlibrary{positioning}
\usetikzlibrary{shapes}
\usetikzlibrary{tikzmark}
\usetikzlibrary{fit}

\author{
    Alexander Thebelt\thanks{Corresponding author} \\
    Imperial College London, \\
    South Kensington, SW7 2AZ, UK. \\
    \texttt{alexander.thebelt18@imperial.ac.uk} \\
    \And
    Jan Kronqvist \\
    Imperial College London, \\
    South Kensington, SW7 2AZ, UK. \\
    \texttt{j.kronqvist@imperial.ac.uk } \\
    \And
    Miten Mistry \\
    Imperial College London, \\
    South Kensington, SW7 2AZ, UK. \\
    \texttt{miten.mistry11@imperial.ac.uk} \\
    \And
    Robert M. Lee \\
    BASF SE, \\
    Ludwigshafen am Rhein, Germany. \\
    \texttt{robert-matthew.lee@basf.com} \\
    \And
    Nathan Sudermann-Merx \\
    BASF SE, \\
    Ludwigshafen am Rhein, Germany. \\
    \texttt{nathan.sudermann-merx@basf.com} \\
    \And
    Ruth Misener \\
    Imperial College London, \\
    South Kensington, SW7 2AZ, UK. \\
    \texttt{r.misener@imperial.ac.uk } \\
}

\begin{document}
    \maketitle

    \doublespacing

    \begin{abstract}
        Gradient boosted trees and other regression tree models perform well in a wide range of real-world,
        industrial applications. These tree models (i) offer insight into important prediction features, (ii)
        effectively manage sparse data, and (iii) have excellent prediction capabilities. Despite their advantages,
        they are generally unpopular for decision-making tasks and black-box optimization, which is due to their
        difficult-to-optimize structure and the lack of a reliable uncertainty measure. \texttt{ENTMOOT} is our new
        framework for integrating (already trained) tree models into larger optimization problems. The contributions
        of \texttt{ENTMOOT} include: (i) explicitly introducing a reliable uncertainty measure that is compatible
        with tree models, (ii) solving the larger optimization problems that incorporate these uncertainty aware tree
        models, (iii) proving that the solutions are globally optimal, i.e.\ no better solution exists. In
        particular, we show how the \texttt{ENTMOOT} approach allows a simple integration of tree models into
        decision-making and black-box optimization, where it proves as a strong competitor to commonly-used frameworks.
    \end{abstract}

    \keywords{Bayesian optimization \and gradient-boosted trees \and deterministic global optimization \and black-box
    optimization}

    \section{Introduction}
    Recently, Bayesian optimization (BO) has become a popular machine learning-based approach for optimizing
    black-box functions, with successful applications including hyperparameter tuning of machine learning algorithms
    \citep{snoek2012hyperparam}, design of engineering systems \citep{forrester2008BayesOpt,mockus1989BayesOpt} and
    drug design \citep{negoescu2011BayesOpt}. A black-box function $f:\mathbb{R}^n \to \mathbb{R}$ can be evaluated
    for $\boldsymbol{x} \in \mathbb{R}^n$ but no further information, such as the derivatives, is available. BO
    proposes sequential function queries $\boldsymbol{x}_{next} \in \mathbb{R}^n$ to iteratively determine the global
    minimizer $\boldsymbol{x^*}$ of black-box function $f(\boldsymbol{x})$ according to Equ.~\eqref{eq:intro_bo_obj}.
    \begin{subequations}
        \label{eq:intro_bo}
        \begin{flalign}
            \boldsymbol{x^*} \in &\underset{\boldsymbol{x} \in \mathbb{R}^n} {\text{argmin}} \; f(\boldsymbol{x}).
            \label{eq:intro_bo_obj} \\
            \; \; \; \text{s.t.} \; \; & g(\boldsymbol{x}) = 0, \label{eq:intro_bo_constr_eq} \\
            & h(\boldsymbol{x}) \leq 0, \label{eq:intro_bo_constr_ineq}
        \end{flalign}
    \end{subequations}
    The input space $\boldsymbol{x}$ may be subject to explicitly known equality constraints $g(\boldsymbol{x})$ and
    inequality constraints $h(\boldsymbol{x})$. BO derives $\boldsymbol{x}_{next}$ based on a data-driven surrogate
    model and an uncertainty measure, quantifying the prediction performance of the surrogate model in different
    search space areas. Both surrogate model prediction and uncertainty measure are combined in an acquisition
    function, capturing the \textit{exploitation / exploration}-trade-off. For detailed surrogate modeling and BO
    reviews, see \citep{brochu2009BayesOpt,frazier2018Tutorial,frazier2016BayesOpt,shahriari2016BO,
        bhosekar2018surrogateReview,mcbride2019overview}. Besides neural networks \citep{snoek2015SBOAnn} and
    Gaussian processes \citep{rasmussen2006GP}, tree-based models, e.g.\ random forests (RFs) \citep{breiman2001RF}
    and gradient-boosted regression trees (GBRTs) \citep{friedman2001GBTGreedyMachine,friedman2002GBTstochastic}, are
    a popular choice for BO surrogate models. Tree-based models give robust prediction performance and can handle
    nonlinear and discontinuous system behavior. Combined with fast and sophisticated training libraries
    \citep{chen2016xgboost,ke2017LightGBM}, tree models scale well to both large data sets and high-dimensional data.
    Moreover, tree-based models naturally support categorical features. \citet{shahriari2016BO} identify unreliable
    uncertainty estimates and the lack of gradients when optimizing tree model-based acquisition functions as the
    main drawbacks, limiting the applicability of tree model-based BO approaches. \par
    \texttt{ENTMOOT} (\textbf{EN}semble \textbf{T}ree \textbf{MO}del \textbf{O}ptimization \textbf{T}ool) is our
    novel framework to handle tree-based models in BO applications. We propose an intuitive distance-based measure to
    provide a reliable uncertainty estimate for tree-based models. \texttt{ENTMOOT} encodes tree-based models,
    uncertainty estimates and applicable constraints as a mathematical program and uses deterministic optimization to
    provide the optimal \textit{exploitation / exploration} trade-off. Using mixed-integer optimization,
    \texttt{ENTMOOT} overcomes the commonly known drawbacks of tree-based models in BO while maintaining their
    advantages. Moreover, it guarantees feasibility of additional constraints (see: Equ.~\eqref{eq:intro_bo}) up to a
    given tolerance where $g(\boldsymbol{x})$ and $h(\boldsymbol{x})$ can be of quadratic, i.e.\ convex and nonconvex
    quadratic, or linear form. \texttt{ENTMOOT} is easy-to-use as it only introduces a maximum of two additonal
    hyperparameters for effective uncertainty control. Our numerical study shows the effectiveness of the proposed
    uncertainty measure and excellent scaling to large and high-dimensional data sets. \texttt{ENTMOOT} is
    competitive with other state-of-the-art black-box optimization libraries, deriving better functional values than
    those found by the competing methods in the majority of runs. \texttt{ENTMOOT} v0.1.4 is open-source and
    available at: \url{pypi.org/project/entmoot} and \url{github.com/cog-imperial/entmoot}.

    \section{Related work} \label{sec:related_work}
    Various successful applications of BO include hyperparameter tuning, chemical and biomedical engineering, and
    aerospace design \citep{frazier2018Tutorial,shahriari2016BO}. BO proposes subsequent black-box queries by
    optimizing an acquisition function that trades-off exploitation and exploration, derived from prediction mean and
    uncertainty quantification of the underlying surrogate model. The most popular model architecture in BO, the
    Gaussian process (GP) \citep{rasmussen2006GP}, is a predictive model with built-in uncertainty considerations
    that has been successfully deployed on various occasions. GPs are flexible and have well-calibrated uncertainty
    quantification making it particularly suitable for BO tasks \citep{jones2001GPpros,osborne2009GPpros}. However,
    the limitations of GPs for large evaluation budgets and high-dimensional search spaces are well-documented.
    Different techniques exploit potential additive structures of black-box functions \citep{gardner2017AdditiveGPs,
        kandasamy2015AdditiveGPs,wang2018BatchedBO}, or map a high-dimensional input feature space to an unknown
    low-dimensional one \citep{garnett2014BOEmbed,nayebi2019BOEmbed,oh2018BOEmbed,wang2016BOBillionDims} to overcome
    these GP limitations. A recent approach by \citet{eriksson2019BOTR} combines local GP models with
    problem-dependent trust regions to allow better scalability for large data set in high-dimensional search spaces.
    \par

    \citet{snoek2015SBOAnn} use \textit{Bayesian neural networks} to tackle BO for large data sets, achieving a
    linear scaling with the number of observations rather than the cubic cost of training a GP. This approach uses
    Bayesian linear regression based on an adaptive set of basis functions which is learned by neural networks.
    \citet{springenberg2016Bohamiann} improved the algorithm by introducing a modified Hamiltonian Monte Carlo method
    . This enhancement offers more robust and scalable behavior compared to standard Bayesian neural networks. We
    include a comparison against the \citet{springenberg2016Bohamiann} implementation, i.e.\ the \texttt{BOHAMIANN}
    algorithm. \par

    Another prominent alternative to GPs are tree-based ensemble methods, e.g.\ RFs \citep{breiman2001RF} and GBRTs
    \citep{friedman2001GBTGreedyMachine,friedman2002GBTstochastic}. Tree-based models consist of decision tree
    ensembles and scale well to large data sets due to parallelization. Fixed-size subsets of available dimensions
    define decision nodes of individual trees to provide robust behavior in high-dimensional search spaces. The good
    predictive performance stems from combing weak tree learners into a tree ensemble. Moreover, tree-based models
    naturally support various data types, and are preferred over GPs when it comes to categorical and conditional
    feature spaces. \citet{hutter2011SequentialModel} first introduced RFs as BO surrogate models. In contrast to
    GPs, RFs do not natively return model uncertainty at test points required to handle exploration in the
    acquisition function. Approaches aiming to estimate RF confidence intervals include the \textit{Jackknife} and
    \textit{Infinitesimal Jackknife} \citep{wager2014Jackknife}. \citet{hutter2011SequentialModel} propose empirical
    uncertainty estimates based on predictions of individual trees which have been shown to work well in practice
    when comparing their algorithm, i.e.\ \texttt{SMAC}, to other state-of-the-art BO frameworks. However,
    \citet{shahriari2016BO} point out common settings for which the uncertainty estimate in \texttt{SMAC} leads to
    undesired behavior, i.e.\ narrow confidence intervals in regions of low data density and uncertainty peaks when
    individual trees strongly disagree. Moreover, \citet{shahriari2016BO} identify the lack of gradients due to the
    discontinuous and non-differentiable response surface of tree-based models as a main drawback ruling out
    gradient-based methods to optimize the acquisition function. As a result, most tree-based BO frameworks rely on
    some sort of sampling approach for optimizing the acquisition function. Another implementation of tree-based
    models in BO is Scikit-Optimize \citep{skopt2018} (\texttt{SKOPT}) supporting GBRTs as an alternative to RFs.
    With GBRTs being boosting-type learners, \texttt{SKOPT} uses uncertainty estimates derived from quantile
    regression \citep{koenker2001QuantileRegression,meinshausen2006QuantileForests}. Multiple GBRT models trained on
    different quantiles of given observations estimate the prediction uncertainty at test points. \texttt{SKOPT} is a
    well-maintained BO library showing competitive performance against other state-of-the-art BO frameworks. Our
    numerical tests compare against \texttt{SMAC3}, the most up-to-date version of \texttt{SMAC}, and both the RF and
    GBRT variants of \texttt{SKOPT}. \par

    Other popular approaches for derivative-free global optimization of black-box functions include evolutionary
    algorithms based on stochastic optimization, e.g.\ \texttt{CMA-ES}, Nelder-Mead (\texttt{NM})
    \citep{nelder1965SimplexMethod} and \texttt{BFGS} \citep{zhu1997BFGS}. We include \texttt{CMA-ES}, \texttt{NM}
    and \texttt{BFGS} as cheap and fast algorithms in our comparison to obtain good baselines for computationally
    expensive approaches. \par
    Design of experiments \citep{garud2017design,hasan2011surrogate} and decision-making using surrogate models is
    enjoying growing popularity in chemical engineering \citep{bhosekar2018surrogateReview}. Some examples use convex
    region linear estimators \citep{zhang2016data}, algebraic equations \citep{boukouvala2017argonaut,
        wilson2017alamo}, artificial neural networks \citep{henao2011surrogateNN,schweidtdmann2019globNN,
        eason2014adaptive,leperi2019110th} and Gaussian processes \citep{palmer2002GP,davis2007kriging,
        caballero2008algorithm,olofsson2019bayesian} as surrogate models with some using global strategies for
    optimization.

        \section{ENTMOOT for design of experiments} \label{sec:entmoot_for_design_of_experiments}
    This section derives the problem formulation representing the core of \texttt{ENTMOOT}.
    Section~\ref{sec:spatial_distance} introduces a distance measure quantifying model uncertainty for tree-based
    surrogate models. Section~\ref{sec:mathematical_programming} presents the full formulation of \texttt{ENTMOOT}
    describing how the exploitation vs.\ exploration trade-off is realized using tree-based models and the proposed
    uncertainty measure for tasks related to BO. We introduce formulations for the squared Euclidean and the
    Manhattan distances as two options to capture model uncertainty.

    \subsection{Capturing model uncertainty via spatial distance}\label{sec:spatial_distance}
    Tree-based models work well as interpolators, i.e.\ they have good prediction performance within the data set and
    especially close to training data. However, they are poor extrapolators with decreasing prediction performance
    for unexplored regions of the input feature space not covered by training data \citep{shahriari2016BO}. Different
    methods estimating tree-based model uncertainty \citep{hutter2011SequentialModel,meinshausen2006QuantileForests,
        wager2014Jackknife} are successful in practice but may cause unintuitive behavior in BO, like
    over-exploration or overconfident uncertainty quantification in regions with sparse data. We propose an intuitive
    estimate $\alpha$ using the distance to the closest point $\boldsymbol{x}_{d}$ in data set $\mathscr{D}$ to
    quantify confidence of model predictions:
    %EQUATION: distance measure
    \begin{equation}
        \label{eq:uncertainty_measure}
        \alpha(\boldsymbol{x}) = \underset{d\in \mathscr{D}}{\text{min}} \lVert \boldsymbol{x} - \boldsymbol{x}_{d}
        \rVert_{p}.
    \end{equation}
    We assume that high values of $\alpha$ indicate large average model errors. Spatial distances are commonly used
    in training algorithms to quantify the model error and in clustering algorithms, e.g.\ k-means
    \citep{lloyd1982KMeans}, to assess the parity of cluster members. Similarly, kernel-based methods like GPs
    \citep{rasmussen2006GP} typically use a distance metric in the kernel covariance function to quantify similarity
    of observations and thus, implicitly, prediction uncertainty. \texttt{ENTMOOT} considers both the squared
    Euclidean and the Manhattan distance metrics, referring to $p=2$ and $p=1$ in Equ.~\eqref{eq:uncertainty_measure}
    . Different distance metrics may be advantageous for different problems. For example, the Manhattan distance is
    usually preferred over the Euclidean distance in high-dimensional settings \citep{aggarwal2001L1L2}. The
    Section~\ref{sec:numerical_studies} provides case studies evaluate how well such a distance-based estimate can
    capture model-uncertainty of tree-based models. For \texttt{ENTMOOT}, GBRTs are the models of choice.

    % example figure showing tree-model mean and uncertainty
    \begin{figure*}
        \begin{center}
            \begin{tikzpicture}
                \node[anchor=west] (main) {
                    \includegraphics[]{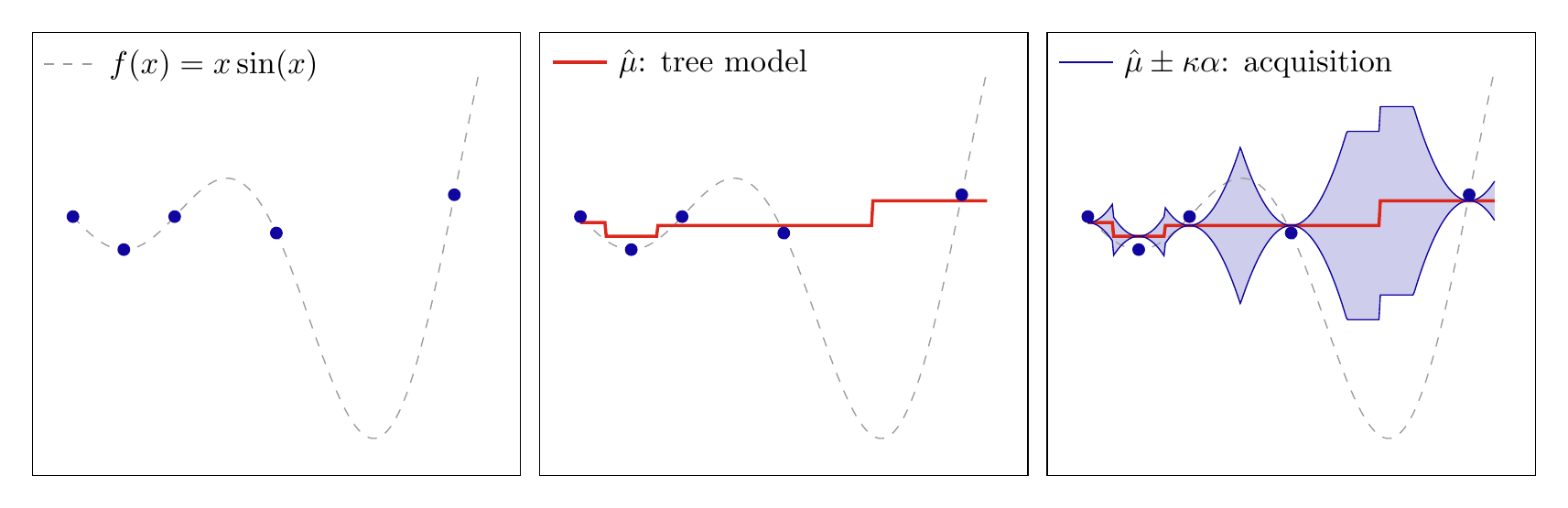}
                };
            \end{tikzpicture}
        \end{center}

        \caption{Illustration of \texttt{ENTMOOT}'s Equ.~\eqref{eq:exp_a} defined exploitation and exploration terms.
        \textbf{(Left)} Plot of the $f(x)=x \sin(x)$ example function and available data points. \textbf{(Middle)}
        GBRT response surface fitted to available data points referring to $\hat{\mu}$ of Equ.~\eqref{eq:exp_a}.
        \textbf{(Right)} Uncertainty intervals representing $\kappa\alpha$ in Equ.~\eqref{eq:exp_a} (lower curve) and
        Equ.~\eqref{eq:penalty_ob} (upper curve) around the mean prediction of $\hat{\mu}$.}
        \label{fig:entmoot_acquisition_function}
    \end{figure*}

    \subsection{Optimizing acquisition functions via mixed-integer optimization} \label{sec:mathematical_programming}
    The general formulation for BO related tasks, e.g.\ design of experiments, is:
    % EQUATIONS: lower confidence bound
    \begin{subequations}
        \label{eq:exploration}
        \begin{flalign}
            \boldsymbol{x}_{next} \in &\underset{\left(\boldsymbol{x}, \boldsymbol{z}, \boldsymbol{y}, \hat{\mu},
            \alpha\right) \in \Omega} {\text{argmin}} \; \hat{\mu} - \kappa \alpha. \label{eq:exp_a} \\
            \; \; \; \text{s.t.} \; \; & \alpha \leq \text{dist}_{d}(\boldsymbol{x}), \; \; \forall d \in
            \mathscr{D}, \label{eq:exp_b}\\
            & 0 \leq \alpha \leq \alpha_{limit},
        \end{flalign}
    \end{subequations}
    where $\Omega$ is a set defined by the constraints in Equ.~\eqref{eq:tree_model} and Equ
    .~\eqref{eq:linking_const} and depending on which uncertainty measure is used Equ.~\eqref{eq:nonconv} or Equ
    .~\eqref{eq:man}. Variable $\hat{\mu} \in \mathbb{R}$ refers to the tree model prediction capturing how Equ
    .~\eqref{eq:exp_a} exploits the underlying surrogate model to find promising areas in the search space. Variables
    $\boldsymbol{z} \in \mathbb{R}^{\mid \mathscr{T} \mid \times \mid \mathscr{L}_t \mid}$ and $\boldsymbol{y} \in \{
    0,1 \}^{n \times \mid \mathscr{V}_t \mid}$ mathematically encode the tree model with the index set of trees
    denoted as $\mathscr{T}$. $\mathscr{L}_t$ and $\mathscr{V}_t$ are dependent on tree $t \in \mathscr{T}$ and
    denote the index sets of leaves and splits, respectively. Section~\ref{sec:encoding_ensemble_tree_models} gives
    more details on the encoding of tree models. Variable $\alpha$ handles exploration and quantifies, according to
    Equ.~\eqref{eq:uncertainty_measure}, the degree of uncertainty expected from prediction $\hat{\mu}$. The
    parameter $\alpha_{limit} \in \mathbb{R}$ provides the upper bound to variable $\alpha$. Variable $\text{dist}_d
    (\boldsymbol{x}) \in \mathbb{R}$ encodes the distance to the closest data point and depends on the distance
    metric used. Both $\alpha_{limit}$ and $\text{dist}_d(\boldsymbol{x})$ are defined in more detail in
    Sections~\ref{sec:encoding_euclidean_distance_squared_uncertainty} and
    \ref{sec:encoding_manhattan_distance_uncertainty} discussing different distance metrics. In BO, Equ
    .~\eqref{eq:exp_a}, or the acquisition function, trades-off exploitation and exploration and its minimizer
    determines the next black-box query point $\boldsymbol{x}_{next}$. The acquisition function incentivizes high
    model uncertainties to force the algorithm to explore unknown search space regions. The positive hyperparameter
    $\kappa \in \mathbb{R}$ balances exploitation and exploration and is a hyperparameter depending on the
    application. The Equ.~\eqref{eq:exp_a} acquisition function, i.e.\ the lower confidence bound (LCB)
    \citep{cox1997LCB}, is available in most BO libraries, e.g.\ \texttt{SMAC3} and \texttt{\texttt{SKOPT}}.
    Figure~\ref{fig:entmoot_acquisition_function} visualizes the interplay of exploitation and exploration as
    implemented in \texttt{ENTMOOT}. When combined with tree-based surrogate models Equ.~\eqref{eq:exp_a} is commonly
    minimized in a stochastic fashion, e.g.\ by random sampling, as the discrete response surface of tree-based
    models precludes effective use of gradient-based optimization methods \citep{shahriari2016BO}. Consequently,
    non-optimal solutions to the acquisition function lead to poorly performing query points $\boldsymbol{x}_{next}$.
    We propose a mixed-integer optimization framework deriving $\epsilon$-global solutions to Equ.~\eqref{eq:exp_a},
    guaranteeing the best trade-off between exploitation and exploration as encapsulated by the acquisition function.
    \par

    \subsubsection{Encoding ensemble tree models} \label{sec:encoding_ensemble_tree_models}
    First, we encode the tree-based model with a mixed-integer formulation initially proposed by
    \citet{misic2017OptimizationEnsembles}. \citet{mistry2018MixedIntegerEmbedded} used the same formulation to
    combine GBRTs together with a continuous regularizer in an optimization problem. Both
    \citet{misic2017OptimizationEnsembles} and \citet{mistry2018MixedIntegerEmbedded}, use mathematical programming
    to determine optimal inputs of tree-based models. We extend this formulation to a full-fledged BO framework with
    a reliable uncertainty measure, where an acquisition function is subsequently optimized to minimize an unknown
    black-box function. The following constraints encode the GBRTs:
    % EQUATIONS: gbrt model
    \begin{subequations}
        \label{eq:tree_model}
        \begin{flalign}
            &\hat{\mu} = \sum\limits_{t\in{\mathscr{T}}} \sum\limits_{l\in{\mathscr{L}_{t}}} F_{t,l} z_{t,l}, & &
            \label{eq:const_a}\\
            &\;\;\;\sum\limits_{l\in{\mathscr{L}_{t}}}\;\;\;\; z_{t,l} = 1, & &\forall t\in \mathscr{T},
            \label{eq:const_b}\\
            &\sum\limits_{\;\;l\in\text{Left\;}_{t,s}}\; z_{t,l} \leq y_{i(s),j(s)}, & &\forall t \in \mathscr{T},
            \forall s \in \mathscr{V}_{t}, \label{eq:const_c}\\
            &\sum\limits_{\;\;l\in\text{Right}_{t,s}} z_{t,l} \leq 1 - y_{i(s),j(s)}, & &\forall t \in \mathscr{T},
            \forall s \in \mathscr{V}_{t}, \label{eq:const_d} \\
            &y_{i,j} \leq y_{i,j+1}, & &\forall i \in \left [ n \right ], \forall j \in \left [ m_i - 1 \right ],
            \label{eq:const_e} \\
            &y_{i,j} \in \{ 0,1  \}, & &\forall i \in \left [ n \right ], \forall j \in \left [ m_i \right ],
            \label{eq:const_f} \\
            &z_{t,l} \geq 0, & &\forall t\in{\mathscr{T}}, \forall l\in{\mathscr{L}_{t}}. \label{eq:const_g}
        \end{flalign}
    \end{subequations}
    We define the tree model prediction $\hat{\mu}$ of Equ.~\eqref{eq:exp_a} in Equ.~\eqref{eq:const_a} as the sum of
    all leaf value parameters $F_{t,l}$. The leaves are indexed by $t\in{\mathscr{T}}$ and $l\in{\mathscr{L}_{t}}$,
    with $\mathscr{T}$ and $\mathscr{L}_t$ defining the set of trees and leaves in every tree $t$, respectively.
    Variables $z_{t,l} \in \mathbb{R}$ function as binary switches, determining which leaves are active. Equ
    .~\eqref{eq:const_b} ensures that only one leaf per tree contributes to the GBRT prediction. Equ
    .~\eqref{eq:const_c}, \eqref{eq:const_d} and \eqref{eq:const_e} force all splits $s \in \mathscr{V}_{t}$, leading
    to an active leaf, to occur in the correct order. Here $\text{Left}_{t,s}$ and $\text{Right}_{t,s}$ denote the
    sets of leaf indices $l \in \mathscr{L}_t$ which are left and right of split $s$ in tree $t$. Binary switches
    $y_{i(s),j(s)}$ determine which splits are active. Optimizing for variables $\boldsymbol{z}$ and $\boldsymbol{y}$
    corresponds to the tree-model contribution in Equ.~\eqref{eq:exp_a}.
    % EQUATIONS: linking
    \begin{subequations}
        \label{eq:linking_const}
        \begin{flalign}
            &x_{i} \geq v_{i,0} + \sum\limits_{j=1}^{m_{i}} \left (v_{i,j} - v_{i,j-1} \right ) \left ( 1 - y_{i,j}
            \right ), & &\forall i \in \left [ n \right ],\\
            &x_{i} \leq v_{i,m_{i}+1} + \sum\limits_{j=1}^{m_{i}} \left (v_{i,j} - v_{i,j-1} \right ) y_{i,j}, &
            &\forall i \in \left [ n \right ],\\
            &x_{i} \in \left [ v_{i}^{L},v_{i}^{U} \right ], & &\forall i \in \left [ n \right ].
        \end{flalign}
    \end{subequations}

\begin{figure*}
        \begin{center}
            \begin{tikzpicture}
                \node[anchor=west] (main) {
                    \includegraphics[width=\textwidth]{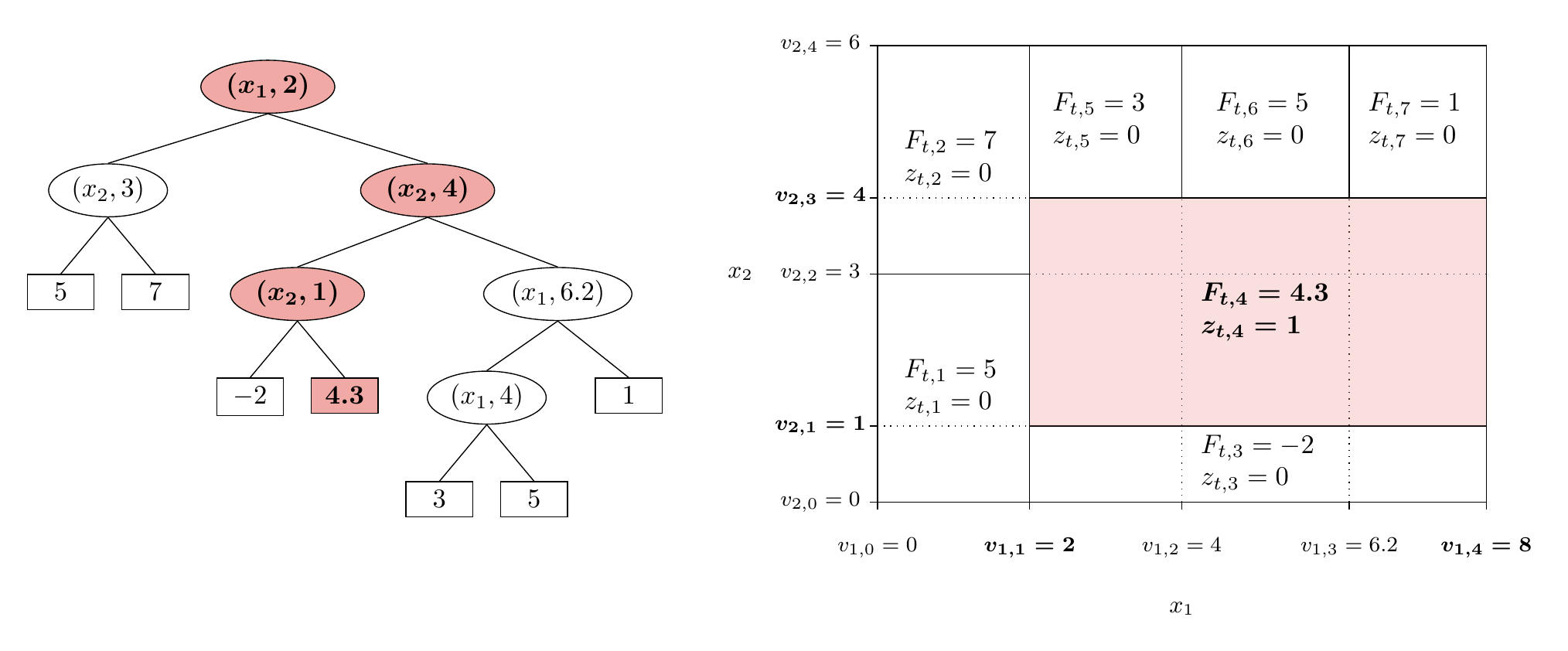}
                };
            \end{tikzpicture}
        \end{center}

        \caption{Example evaluation of a decision tree $t$ of a tree ensemble $\mathscr{T}$ for $\boldsymbol{x}=(4.2,
        2.8)^T$. \textbf{(Left)} Gradient-boosted tree trained for two dimensions. \textbf{(Right)} Domain partition
        according to tree model formulation in Equ.~\eqref{eq:tree_model} and Equ.~\eqref{eq:linking_const}.}
        \label{fig:tree_model}
    \end{figure*}

    We introduce linking constraints according to Equ.~\eqref{eq:linking_const} to assign separate intervals $v_{i,
    j}$ defined by the tree model splits back to the original continuous search space $\boldsymbol{x} \in
    \mathbb{R}^n$. Splits are ordered based on their numerical value according to $v_i^L = v_{i,0} < v_{i,1} < ... <
    v_{i,m_i} < v_{i,m_i+1} = v_i^U$ with $v_i^L$ and $v_i^U$ denoting upper and lower bound, respectively.
    Figure~\ref{fig:tree_model} shows a simple example of how a tree $t$ of tree ensemble $\mathscr{T}$ partitions a
    two-dimensional feature space according to Equ.~\eqref{eq:tree_model} and Equ.~\eqref{eq:linking_const}.

    % figure showing tree model workings

    \subsubsection{Encoding Euclidean distance squared uncertainty}
    \label{sec:encoding_euclidean_distance_squared_uncertainty}
    We derive the exploration term $\alpha$ similar to the Section~\ref{sec:spatial_distance} uncertainty measure,
    considering the Euclidean distance squared to the closest data point of data set $\mathscr{D}$. Constraints
    \eqref{eq:nonconv} complete Equ.~\eqref{eq:exploration} by defining $\text{dist}_d(\boldsymbol{x})$:
    % EQUATIONS: squared euclidean distance
    \begin{subequations}
        \label{eq:nonconv}
        \begin{flalign}
            & \text{dist}_d({\boldsymbol{x})} \leq \lVert \boldsymbol{\sigma}^{-1}_{\mathscr{D},diag} (\boldsymbol{x}
            - \boldsymbol{\mu}_\mathscr{D}) - \boldsymbol{x}_d \rVert_{2}^{2}, & &\forall d\in \mathscr{D},
            \label{eq:nonconv_a}\\
            &\alpha_{limit} = \zeta \ \text{Var}(\boldsymbol{y}_{\mathscr{D}}). \label{eq:nonconv_b} & &
        \end{flalign}
    \end{subequations}
    Equ.~\eqref{eq:exp_a} incentivizes larger values for $\alpha$, i.e.\ greater distance to the nearest data
    point, while Equ.~\eqref{eq:nonconv_a} distance constraints to other data points automatically
    become redundant. As the exploration term $\alpha$ grows
    quadratically, Equ.~\eqref{eq:nonconv_b} enforces a limit $\alpha_{limit}$ to restrict exploration. The product
    of hyperparmeter $\zeta \in \mathbb{R}$ and the variance observed in data set target values
    $\boldsymbol{y}_{\mathscr{D}}$ has shown, in our preliminary studies, to be a good measure. Equ
    .~\eqref{eq:nonconv} define our bounded data distance measure. An intuitive interpretation of bounding the data
    distance is that the model uncertainty is bounded. In other words, setting the hyperparameter $\zeta$
    incorporates how much more or less variance we expect compared to what is already evident in data set
    $\mathscr{D}$. Parameters $\boldsymbol{\sigma}_{\mathscr{D},diag} \in \mathbb{R}^{n \times n}$ and
    $\boldsymbol{\mu}_\mathscr{D} \in \mathbb{R}^{n}$ denote standard deviation and mean of dataset $\mathscr{D}$ and
    scale variables $\boldsymbol{x}$ as $\boldsymbol{x}_d$ is given in standardized form. The optimization problem
    defined by objective \eqref{eq:exp_a} and constraints \eqref{eq:tree_model}, \eqref{eq:linking_const} and
    \eqref{eq:nonconv} is a nonconvex mixed-integer quadratic problem.

    The nonconvexity arises through the Equ.~\eqref{eq:exp_a} negative contribution of variable $\alpha$ in
    combination with the quadratic term in Equ.~\eqref{eq:nonconv_a}. This corresponds to
    minimizing the blue lower
    curve on the right-hand side of Fig.~\ref{fig:entmoot_acquisition_function} which is described by piecewise
    concave quadratics centered at data points. The resulting optimization problem can be solved to $\epsilon$-global
    optimality with commercial software, e.g.\ Gurobi~9 \citep{gurobi}.

    \subsubsection{Encoding Manhattan distance uncertainty} \label{sec:encoding_manhattan_distance_uncertainty}
    To encode an uncertainty measure using the Manhattan distance metric, we propose a formulation similar to
    \citet{giloni2002L1}. The following constraints are added to the optimization problem stated in Equ
    .~\eqref{eq:exploration}:
    % EQUATIONS: manhattan distance
    \begin{subequations}
        \label{eq:man}
        \begin{flalign}
            & \text{dist}_d(\boldsymbol{x}) \leq \sum\limits_{i \in \left [ n \right ]} r_i^{d,+} + r_i^{d,-}, &
            &\forall d\in \mathscr{D}, \label{eq:man_a}\\
            &x_i^{d} - \sigma_i^{-1}(x_{i} - \mu_i) = r_i^{d,+} - r_i^{d,-},& &\forall d\in \mathscr{D}, \forall i
            \in \left [ n \right ],\label{eq:man_b}\\
            &r_i^{d,+} \cdot r_i^{d,-} = 0,& &\forall d\in \mathscr{D}, \forall i \in \left [ n \right ],
            \label{eq:man_c}\\
            &r_i^{d,+}, r_i^{d,-} \geq 0, & &\forall d\in \mathscr{D}, \forall i \in \left [ n \right ],
            \label{eq:man_d}\\
            &\alpha_{limit} = \zeta \ \text{Var}(\boldsymbol{y}_{\mathscr{D}}). \label{eq:man_e}
        \end{flalign}
    \end{subequations}
    Depending on the left-hand side of Equ.~\eqref{eq:man_b} being positive or negative, auxiliary variables $r_i^{d,
    +} \in \mathbb{R}$ or $r_i^{d,-} \in \mathbb{R}$ take the positive distance to the nearest data point in
    dimension $i$, respectively. Equ.~\eqref{eq:man_c} forces either $r_i^{d,+}$ or $r_i^{d,-}$ to be zero thereby
    leading to $\alpha$ taking the value of the Manhattan distance to the closest data point in Equ.~\eqref{eq:man_a}
    . Parameters $\sigma_i$ and $\mu_i$ correspond to entries $(i,i)$ of $\boldsymbol{\sigma}_{\mathscr{D},diag}$ and
    $i$ of $\boldsymbol{\mu}_\mathscr{D}$, respectively. The optimization problem defined by objective
    \eqref{eq:exp_a} and constraints \eqref{eq:tree_model}, \eqref{eq:linking_const} and \eqref{eq:man} is a
    mixed-integer linear problem. Bilinear terms in Equ.~\eqref{eq:man_c} are implemented as special ordered sets
    \citep{tomlin1988SOS} and the problem can be solved to $\epsilon$-global optimality using Gurobi~9. \par
    Compared to the Section~\ref{sec:encoding_euclidean_distance_squared_uncertainty} squared Euclidean distance
    metric, the formulation in Equ.~\eqref{eq:man} requires a significantly higher number of constraints. The Equ
    .~\eqref{eq:nonconv} squared Euclidean distance constraints add one constraint per data point $d \in \mathscr{D}$
    . In contrast, Equ.~\eqref{sec:encoding_manhattan_distance_uncertainty} requires one constraint per dimension $i
    \in \left[ n\right]$ per data point $d \in \mathscr{D}$. This difference becomes noticeable for high-dimensional
    problems and translates to longer run times for the Manhattan distance formulation as can be seen in
    Table~\ref{tab:runtime}.

    \section{ENTMOOT for decision-making under uncertainty} \label{sec:decision_making_under_uncertainty}
    This section builds on Section~\ref{sec:entmoot_for_design_of_experiments} by deriving a formulation for a
    related challenge, i.e.\ decision-making under uncertainty. This considers the case where enough data was
    gathered to build reliable tree-based models which are then incorporated in wider decision-making problems. A
    similar formulation considering the previously introduced uncertainty measures is used to incentivize solutions
    close to training data, i.e.\ existing knowledge, to minimize the risk of model failure. Moreover, we present
    algorithmic modifications to allow scalability for handling applications with large data sets and tree models. \par
    To stay close to observed data points in order to minimize the risk of model failure, i.e.\ predictions with high
    errors, we consider model uncertainty as a penalty:
    \begin{subequations}
        \label{eq:penalty}
        \begin{flalign}
            \boldsymbol{x}_{next} \in &\underset{\left(\boldsymbol{x}, \boldsymbol{z}, \boldsymbol{y}, \hat{\mu},
            \alpha\right) \in \Omega} {\text{argmin}} \; \hat{\mu} + \kappa \alpha. \label{eq:penalty_ob} \\
            \; \; \; \text{s.t.} \; \; & \text{dist}_{d}(\boldsymbol{x}) \leq \alpha + M(1-b_d), \; \; \forall d \in
            \mathscr{D}, \label{eq:bigm_b}\\
            &\sum\limits_{d\in \mathscr{D}} b_d = 1, \label{eq:bigm_c}\\
            &b_d \in \{0,1\}, \; \; \forall d \in \mathscr{D},\\
            & \alpha \geq 0.
        \end{flalign}
    \end{subequations}
    When uncertainty contributes as a penalty to the objective function, ``big-M'' constraints
    \citep{nemhauser1988BigM} are required to only consider the closest data point. Binary variables $b_{d} \in \{0,
    1\}$ are included into the set of optimization variables and function as a binary switch. When $b_d = 0$, a
    sufficiently large $M \in \mathbb{R}$ makes Equ.~\eqref{eq:bigm_b} constraints redundant and thus, deactivates
    data points that are not the closest ones to $\boldsymbol{x}$. For $b_d = 1$, coefficient $M$ is multiplied by
    $0$ and disappears, causing variable $\alpha$ to take the value of Equ.~\eqref{eq:bigm_b} left-hand side
    $\text{dist}_d(\boldsymbol{x})$ which depends on what distance metric is used. Equation~\eqref{eq:bigm_c}
    enforces exactly one active data point. For this formulation, a sufficiently large $M$ is:
    \begin{equation}
        M = \sum\limits_{i \in \left [ n \right ]} \big\lVert \sigma_i^{-1}(v_i^U - v_i^L) \big\rVert_{p}
    \end{equation}
    The optimization problem defined by objective \eqref{eq:penalty_ob} and constraints \eqref{eq:tree_model},
    \eqref{eq:linking_const} and \eqref{eq:nonconv} is a convex mixed-integer quadratic problem. It corresponds to
    minimizing the blue upper curve on the right-hand side of Fig.~\ref{fig:entmoot_acquisition_function} which is
    described by piecewise convex quadratics centered at data points. When choosing the Equ.~\eqref{eq:man} Manhattan
    distance to define $\text{dist}_{d}(\boldsymbol{x})$ the formulation becomes a mixed-integer linear problem. Both
    problems can be solved to $\epsilon$-global optimality with commercial software, e.g.\ Gurobi~9 \citep{gurobi}.

    \subsubsection{Handling large data sets} \label{sec:handling_large_data}
    The Section~\ref{sec:decision_making_under_uncertainty} formulation given in Equ.~\ref{eq:penalty} introduces
    individual constraints for every data point $d \in \mathscr{D}$. The optimization problem grows quickly with
    increasing amounts of data. As a solution we propose using a clustering algorithm, e.g.\ k-means
    \citep{lloyd1982KMeans}, as a pre-processing step. The formulation given in Equ.~\ref{eq:penalty} is replaced by:
    \begin{subequations}
        \label{eq:cpenalty}
        \begin{flalign}
            \boldsymbol{x}_{next} \in &\underset{\left(\boldsymbol{x}, \boldsymbol{z}, \boldsymbol{y}, \hat{\mu},
            \alpha\right) \in \Omega} {\text{argmin}} \; \hat{\mu} + \kappa \alpha. \\
            \; \; \; \text{s.t.} \; \; & \text{dist}_{k}(\boldsymbol{x}) \leq \alpha + M(1-b_k), \; \; \forall k \in
            \mathscr{K}, \label{eq:cbigm_b}\\
            &\sum\limits_{k\in \mathscr{D}} b_k = 1, \label{eq:cbigm_c}\\
            &b_k \in \{0,1\}, \\
            & \alpha \geq 0.
        \end{flalign}
    \end{subequations}
    $\mathscr{K}$ denotes the set of cluster centers derived from the clustering. Model uncertainty is then
    quantified as the distance $\text{dist}_{k}(\boldsymbol{x})$ to cluster centers instead of data points. For large
    data sets, this may also lead to more robust behavior of the algorithm as noisy observations will likely be
    grouped into clusters represented by a single cluster center derived from the mean of all members.

    \subsubsection{Handling large tree models}
    \label{sec:handling_large_tree}
\begin{figure}
        \begin{center}
            \begin{tikzpicture}
                \node[anchor=west] (main) {
                    \includegraphics[trim={4.0cm 6.28cm 3.6cm 6cm},clip,width=0.45\paperwidth]{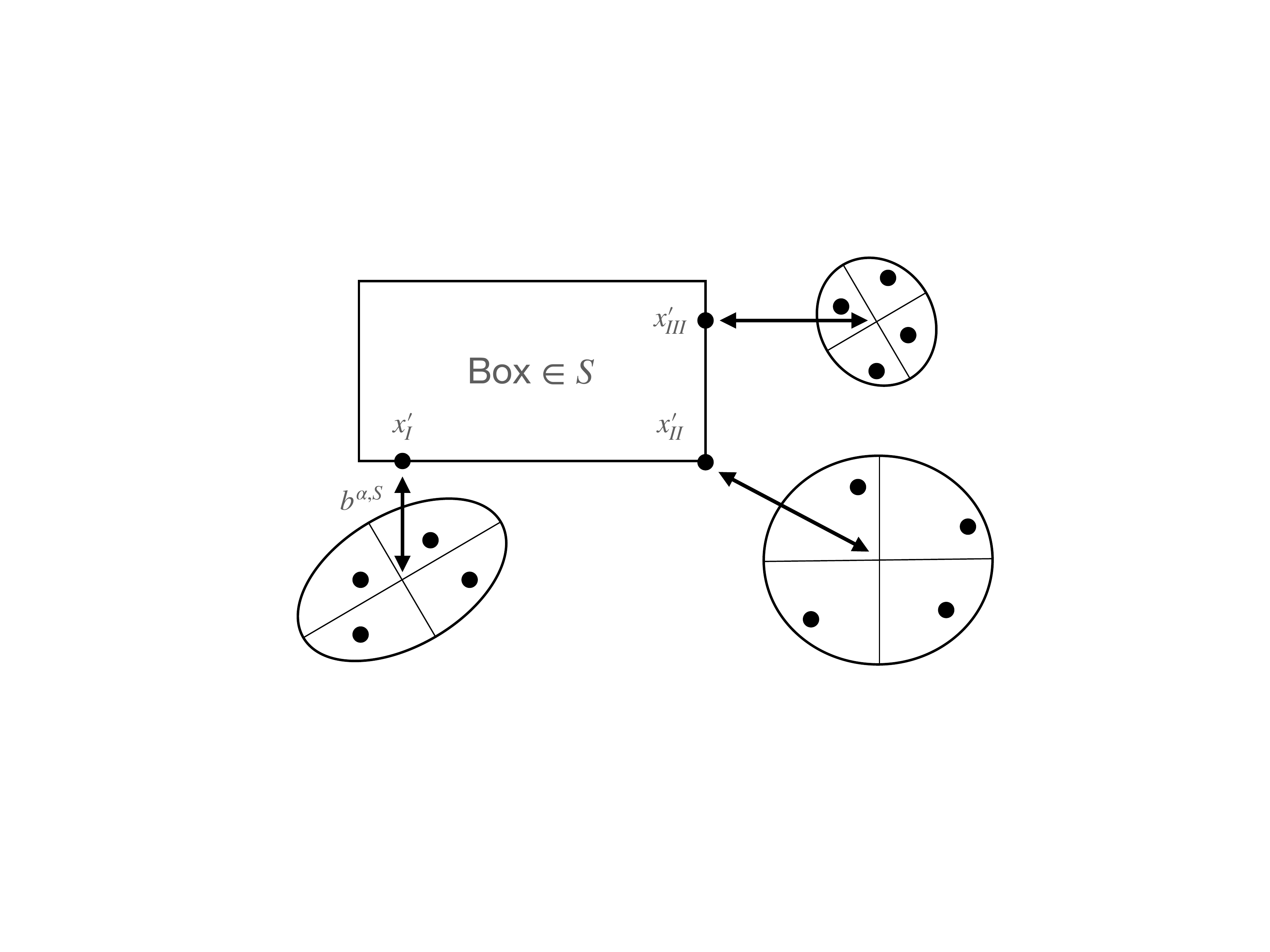}
                };
            \end{tikzpicture}
        \end{center}
        \caption{Explicit evaluation of Euclidean distance squared to cluster center projections onto box $S$ for
        effective computation of $b^{\alpha,S}$.}
        \label{fig:cluster}
    \end{figure}

    When GBRT models become extremely large, i.e.\ more than 2000 trees with a large number of splits per tree,
    Gurobi~9 struggles to prove optimality for the large-scale MINLP. To handle these large-scale instances, we
    propose a more effective bounding strategy \citep{mistry2018MixedIntegerEmbedded} in the branch-and-bound
    algorithm used to solve the underlying mixed-integer problem. \\
    The specifically tailored branch-and-bound approach uses spatial branching over the domain $[\boldsymbol{v}^{L},
    \boldsymbol{v}^{U}]$. Branch-and-bound characterizes every sub-domain in a minimization problem by a lower
    objective bound on the best possible solution in the domain. Individual domains are rejected for infeasible
    subproblems or when their lower bound exceeds the current best feasible solution, i.e.\ the upper bound.
    Branch-and-bound thereby aims to avoid explicit enumeration of all possible solutions \citep{land1960Autom,
        morrison2016BAB}. Like \citet{mistry2018MixedIntegerEmbedded}, we use strong branching to reduce the search
    space. To compute a new lower bound $\hat{R}^{S}$ for domain $S$, \texttt{ENTMOOT} decomposes the
    Objective~\eqref{eq:exp_a} into two parts:
    \begin{equation}
        \label{eq:lowerbound}
        \hat{R}^{S} = b^{\hat{\mu},S} + b^{\alpha,S},
    \end{equation}
    where $b^{\hat{\mu},S}$ and $b^{\alpha,S}$ define the objective lower bounds in domain $S$ for the GBRT model and
    penalty function, respectively. Computing the tightest objective lower bound for $\hat{\mu}$ in domain $S$ is
    NP-hard \cite{misic2017OptimizationEnsembles} and hence, difficult to provide for large GBRT model instances. The
    approach presented here derives a weaker objective lower bound $b^{\hat{\mu},S}$ by \textit{partition refinement}
    \citep{mistry2018MixedIntegerEmbedded}.
    We compute $b^{\alpha,S}$ in time linear to the number of clusters multiplied by the number of dimensions, i.e.\
    $\mathcal{O}(\lvert \mathscr{K} \rvert \cdot \ n)$.
    This is done by calculating the distance from cluster center $\boldsymbol{x}_k$ to the projection
    $\boldsymbol{x}'_k$ of the cluster center on box $S$. The projection $x'_k=(x'_{k,1}, ..., x'_{k,n})^T$ has its
    elements $x_{k,i}'$ defined as:
    % equations for x' projections
    \begin{equation}
        \label{eq:x_proj}
        x_{k,i}'=
        \begin{cases}
            x_{k,i},& \text{if } x_{k,i} \in \left[ v_i^{L,S},v_i^{U,S} \right]\\
            v_i^{U,S},& \text{if } x_{k,i} \geq v_i^{U,S} \\
            v_i^{L,S},& \text{if } x_{k,i} \leq v_i^{L,S}
        \end{cases}
        ,
    \end{equation}
    with $\boldsymbol{v^{L,S}}$ and $\boldsymbol{v^{U,S}}$ denoting upper and lower bounds of box $S$, respectively.
    The algorithms picks the minimum of these distances for $b^{\alpha,S}$ or sets it to 0 in case one of the cluster
    centers is contained in domain $S$. Figure~\ref{fig:cluster}  depicts this procedure. In large data settings the
    number of clusters can be significantly smaller than the number of data points, i.e.\ $\lvert \mathscr{K} \rvert
    << n$. As $\alpha$ increases for regions distant from training data, deriving a weak bound for $\hat{\mu}$  can
    often reject large domains as the weak lower bound surpasses the current best feasible solution. For domains not
    rejected based on this penalty condition, $b^{\hat{\mu},S}$ is recomputed to derive tighter bounds for
    $\hat{\mu}$ in domain $S$. After every iteration, the best feasible solution and the lowest lower bound converge
    and ultimately prove global optimality. In practice, the algorithm terminates after reaching a pre-defined
    optimality gap between best feasible solution and smallest lower bound. From the structure of the problem, we
    know that pre-defined cluster centers by definition have $\alpha(\boldsymbol{x}_{k})=0$. A good initial feasible
    solution $\boldsymbol{x}_{\textit{feas}}$ can therefore be derived by:
    \begin{equation}
        \label{eq:feas}
        \boldsymbol{x}_{\textit{feas}} = \underset{k\in \mathscr{K}}{\text{argmin}} \; \left \{ \hat{\mu}
        (\boldsymbol{x}_{k}) \right \}.
    \end{equation}
    In Section~\ref{sec:numerical_studies} we present results using a proof-of-concept implementation that shows how
    the specifically tailored branch-and-bound algorithm with proposed improvements from
    Section~\ref{sec:handling_large_tree} and Section~\ref{sec:handling_large_data} leads to faster lower bounding
    for large tree-model instances.

    \section{The ENTMOOT software package}
    \begin{figure*}
        \begin{center}
            \begin{tikzpicture}
                \node[anchor=west] (main) {
                    \includegraphics[]{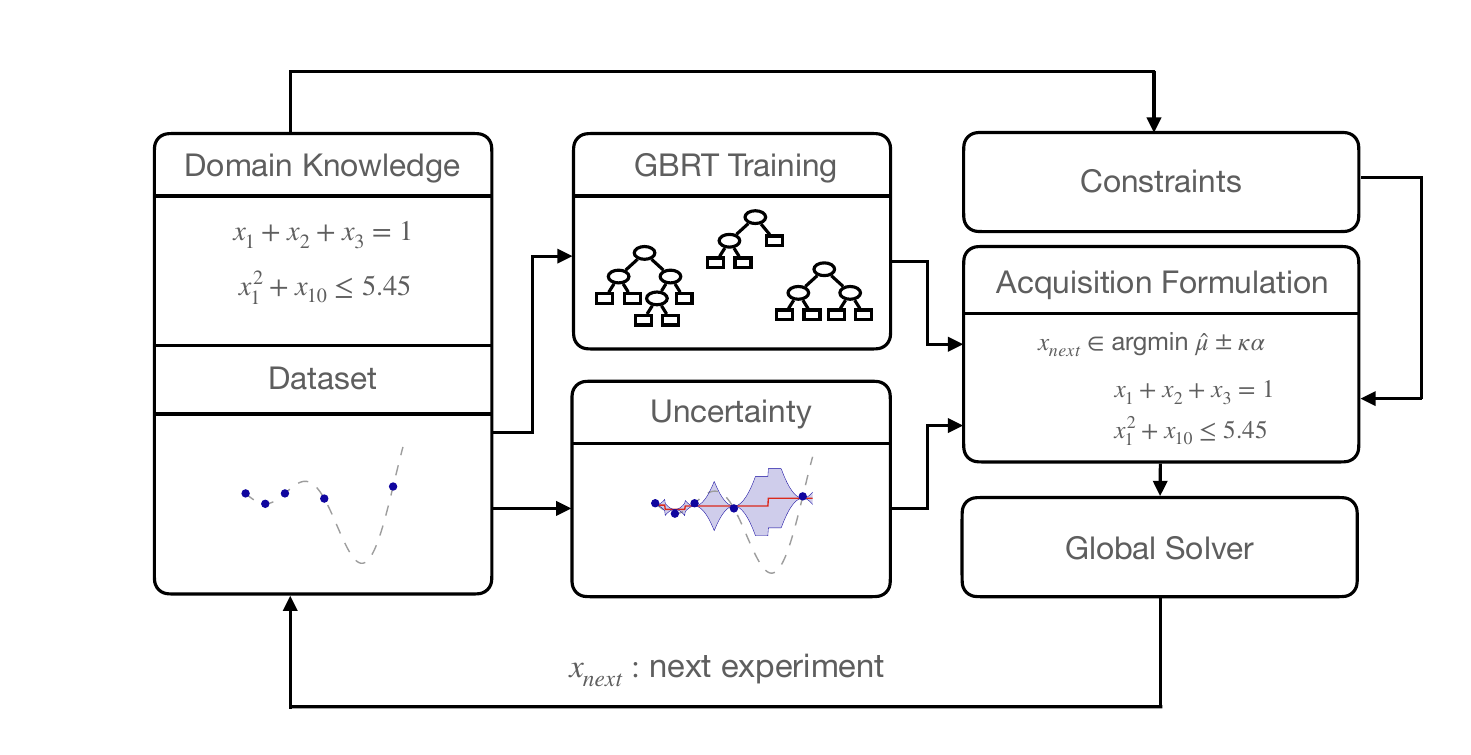}
                };
                \node[anchor=south, yshift=0.0cm] at (main.north) (logo) {
                    \includegraphics[width=0.3\paperwidth]{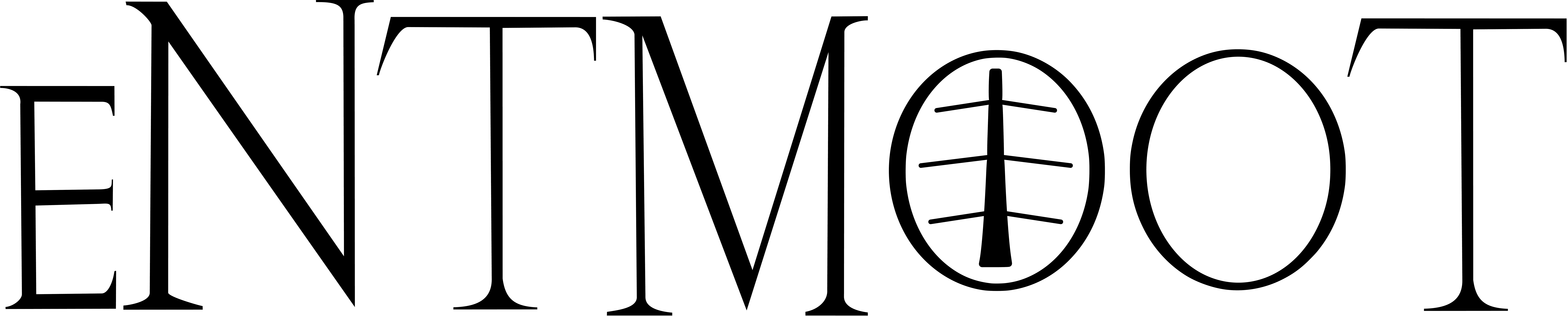}
                };

                \node[fit= (main) (logo),line width=1.2pt,minimum width=0.7\paperwidth, draw] (box) {};
            \end{tikzpicture}
        \end{center}
        \caption{Overview of the \texttt{ENTMOOT} toolbox. Available at: \protect\url{pypi.org/project/entmoot} and
        \protect\url{github.com/cog-imperial/entmoot}}
    \end{figure*} \label{fig:entmoot}

    \texttt{ENTMOOT}'s functionality is summarized in Figure~\ref{fig:entmoot}. The tool processes data input
    provided by the user and hands it over to the global solver to derive $x_{next}$. The user can provide initial
    data and application specific constraints regarding features $\boldsymbol{x}$. The data is used to train GBRT
    models using LightGBM \citep{ke2017LightGBM} with hyperparameters provided by the user. Moreover, the user
    specifies which distance metric is used for uncertainty quantification. Both squared Euclidean distance (see:
    Section~\ref{sec:encoding_euclidean_distance_squared_uncertainty}) and Manhattan distance (see:
    Section~\ref{sec:encoding_manhattan_distance_uncertainty}) are available. Moreover, uncertainty can be quantified
    with respect to individual data points or cluster centers (See Section~\ref{sec:handling_large_data} derived from
    a k-means run. \par
    The acquisition function combines both tree model and model uncertainty. The user can get suggestions $x_{next}$
    far from training data as explained in Section~\ref{sec:entmoot_for_design_of_experiments} or close to existing
    knowledge according to Section~\ref{sec:decision_making_under_uncertainty}. The problem formulation is combined
    with additional user constraints and given to the global solver. \texttt{ENTMOOT} uses the user-specified
    black-box evaluation budget to terminate the sequential optimization loop. \texttt{ENTMOOT} is available on
    \url{pypi.org/project/entmoot} and \url{github.com/cog-imperial/entmoot}.

    \section{Numerical studies} \label{sec:numerical_studies}
    This section evaluates \texttt{ENTMOOT}'s performance on a variety of black-box optimization tasks: a 60D rover
    navigation problem \citep{wang2018BatchedBO}, an 80D fermentation optimization \citep{znad2004KineticModel,
        elqotbi2013CFDModel}, a 14D robot pushing task \citep{wang2018BatchedBO} and various multidimensional
    synthetic benchmark functions. All problems are challenging and commonly used for benchmarking of global
    black-box optimization tools. We compare against a competitive set of state-of-the-art black-box optimization
    algorithms from different fields of literature: \texttt{BOHAMIANN} \citep{springenberg2016Bohamiann},
    \texttt{SMAC3} \citep{hutter2011SequentialModel}, \texttt{SKOPT} \citep{skopt2018}, \text{CMA-ES}
    \citep{hansen2006CMA}, \texttt{BFGS} \citep{zhu1997BFGS}, \texttt{NELDER-MEAD} \citep{nelder1965SimplexMethod}.
    \texttt{DUMMY} refers to random search and is provided as a weak baseline. For \texttt{SKOPT}, we compare against
    three different variants which use different underlying surrogate models: \texttt{SKOPT-GBRT} with
    gradient-boosted trees, \texttt{SKOPT-RF} with random forests and \texttt{SKOPT-GP} with Gaussian processes. For
    \texttt{ENTMOOT}, we test both metrics for the distance-based uncertainty estimate as proposed in
    Section~\ref{sec:encoding_euclidean_distance_squared_uncertainty} and
    Section~\ref{sec:encoding_manhattan_distance_uncertainty}, with \texttt{ENTMOOT} and \texttt{ENTMOOT-L1}
    referring to using the squared Euclidean distance and Manhattan distance, respectively. The label
    \texttt{ENTMOOT-LARGE} refers to using a larger GBRT ensemble, i.e.\ ensembles using a larger number of tree
    estimators. While \texttt{ENTMOOT} with a smaller number of trees is easier to optimize due to the smaller
    resulting optimization model, \texttt{ENTMOOT-LARGE} may have better prediction capabilities as the underlying
    model is significantly larger. Including both modes in the numerical studies allows us to analyze this trade-off.
    Hyperparameters for the training of \texttt{ENTMOOT}, \texttt{ENTMOOT-L1} and \texttt{ENTMOOT-LARGE} are given in
    Appendix~\ref{app:entmoot_details}. Reasonable values for hyperparameters were picked and
    parameter settings are constant for each mode throughout all tests to allow a fair comparison with other
    approaches tested. If not indicated otherwise, all algorithms use default settings without hyperparameter tuning
    for individual runs. Each method is given 50 initial points which are randomly generated from benchmark functions based on random
    seeds, i.e.\ the same random seeds are used for every method and are reported in Appendices~\ref{app:d2},
    \ref{app:d3} and \ref{app:d4}. Each run has a total budget of 300 black-box
    evaluation. We report median and confidence intervals by taking the \nth{1} and \nth{3} quartile
    of the vector of best objective values found for all random seeds at every iteration. The range
    between minimum and maximum of the best objective values found is divided by four to determine thresholds
    for the quartiles. The same random seeds are also used to make optimization methods and noisy black-box
    functions, i.e.\ rover navigation and robot push, reproducible. The appendices provide all
    random states and more details on the individual algorithms and benchmark functions. \par
    This section is divided into four subsections to show: (i) the capabilities of the proposed uncertainty measure
    independent of the implementation (Section~\ref{sec:uncertainty_study}); (ii) how our proposals contribute to
    \texttt{ENTMOOT}'s competitive performance against other tree-based black-box optimization algorithms
    (Section~\ref{sec:tree_study}), (iii) how \texttt{ENTMOOT} compares against other state-of-the-art frameworks on
    commonly-used black-box optimization benchmarks (Section~\ref{sec:perf_study}) and (iv) how algorithmic
    extensions of \texttt{ENTMOOT} can be used for large-scale applications (Section~\ref{sec:large_study}).

    % uncertainty study
    \begin{figure*}
        \begin{center}
            \begin{tikzpicture}
                \node[anchor=west] (unc_err) {
                    \includegraphics[]{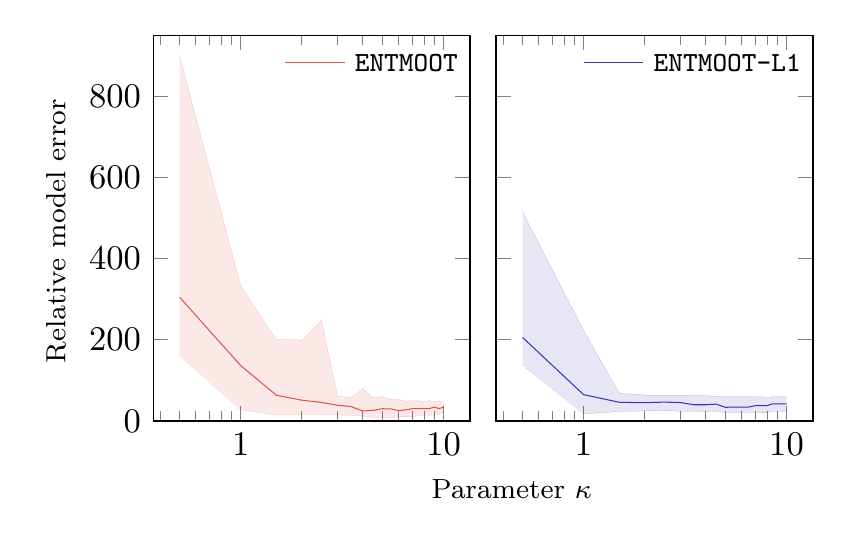}
                };

                \node[anchor=west, xshift=-0.5cm] at (unc_err.east) (unc_pred) {
                    \includegraphics[]{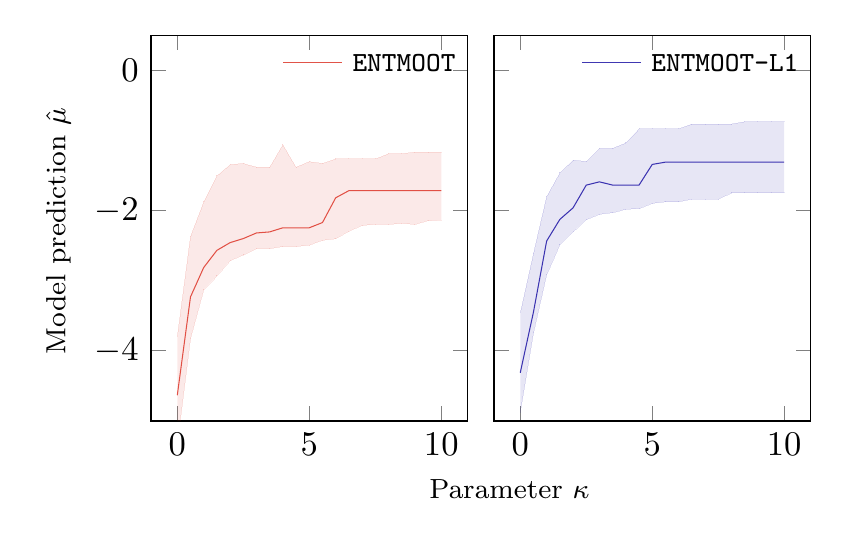}
                };
            \end{tikzpicture}
        \end{center}
        \caption{Analysis of uncertainty measure as described in Section~\ref{sec:uncertainty_study} using proposed
        optimal points. Hyperparameter $\kappa$ controls proximity of $\boldsymbol{x}_{next}$ to training data, i.e.\
            high values of $\kappa$ correspond to closer proximity. \textbf{(Left)} Relative error of tree-based
            models for squared Euclidean distance. \textbf{(Left-Middle)} Relative error of tree-based models for
            Manhattan distance. \textbf{(Middle-Right)} Optimal tree model prediction for squared Euclidean distance.
            \textbf{(Right)} Optimal tree model prediction for Manhattan distance.}
        \label{fig:entmoot_uncertainty_study}
    \end{figure*}

    \subsection{Data distance as an effective measure for uncertainty} \label{sec:uncertainty_study}
    Section~\ref{sec:spatial_distance} proposes using spatial distance as a measure to estimate model uncertainty for
    tree-based models. While distance metrics are commonly used to describe model errors in training algorithms,
    there is still a need to analyze how well they capture model uncertainty in tree-based models. With larger
    distances to training data we expect worse tree model prediction performance. We test this hypothesis by changing
    the objective according to Equ.~\eqref{eq:penalty}:
    % EQUATIONS: lower confidence bound
    \begin{equation}
        \label{eq:objective_alt}
        \boldsymbol{x}_{next} \in \underset{\left(\boldsymbol{x}, \boldsymbol{z}, \boldsymbol{y}, \hat{\mu},
        \alpha\right) \in \Phi} {\text{argmin}} \; \hat{\mu} + \kappa \alpha,
    \end{equation}
    where $\Phi$ is a set defined by the constraints in Equ.~\eqref{eq:tree_model}, Equ.~\eqref{eq:linking_const} and
    Equ.~\eqref{eq:penalty}. Changing the sign in Equ.~\eqref{eq:objective_alt} leads to a positive contribution of
    the distance-based uncertainty measure. This forces \texttt{ENTMOOT} to stay close to training data instead of
    incentivizing exploration. Adjusting $\kappa$ in Equ.~\eqref{eq:objective_alt} controls how close the optimal
    solution is to training data. For the modified optimization model, we expect increasing tree model errors at
    optimal solutions with decreasing values of $\kappa$. \par
    We test on the rover trajectory planning problem which was initially proposed by \citet{wang2018BatchedBO}. The
    goal is to propose a fixed number of points in a 2D-plane determining the trajectory of a rover. Different
    terrain properties influence the reward associated with each proposal. This study uses a smaller 20D variant of
    the rover trajectory planning problem. Fig.~\ref{fig:entmoot_uncertainty_study} shows the results of the study
    when adding Equ.~\eqref{eq:nonconv} or Equ.~\eqref{eq:man} to set $\Phi$ to define the distance metric. As
    expected, large values for hyperparameter $\kappa$ lead to less error in the tree model prediction. Moreover, the
    optimal tree model prediction increases with growing values for $\kappa$ as the Equ.~\eqref{eq:objective_alt}
    acquisition function becomes increasingly dominated by the distance measure functioning as a regularizer. The
    flattening of all curves presented in Fig.~\ref{fig:entmoot_uncertainty_study} corresponds to \texttt{ENTMOOT}
    converging to the best previously observed data point. Based on this analysis, we can use distance to data points
    as a measure for uncertainty as it successfully captures prediction performance of tree-based models. \par
    This uncertainty metric is flexible and easily extendable if more information about the black-box function
    becomes available. For example, if we have information about the input sensitivity, then uncertainty metric terms
    could be multiplied with pre-determined coefficients to weight the uncertainty contribution of individual features.
    Similarly, we can incorporate output sensitivity, e.g.\protect\ assign larger weights to features for which output
    sensitivity is high. Alternatively, we could use a tree-based uncertainty metric, e.g.\protect\ as proposed by
    \citet{misic2017OptimizationEnsembles}.
    % tree study
    \begin{figure*}
        \begin{center}
            \begin{tikzpicture}
                \node[anchor=north west, yshift=3.2cm, xshift=0.5cm] (label_ros20) {\textbf{Rosenbrock 20D}};
                \node[anchor=west, xshift=0.08cm] (ros20) {
                    \includegraphics[]{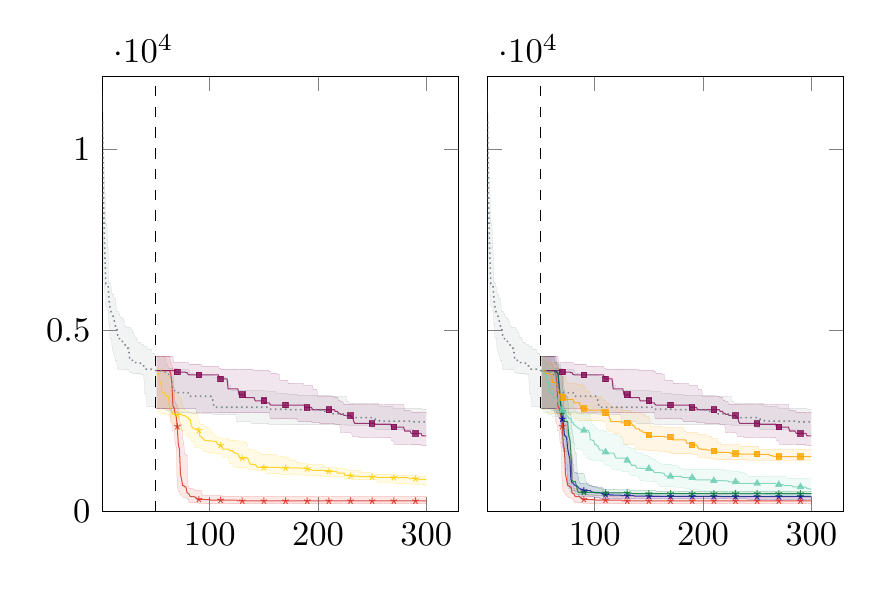}
                };

                \node[anchor=north west, yshift=3.2cm, xshift=9.35cm] (label_ros40) {\textbf{Rosenbrock 40D}};
                \node[anchor=west, xshift=8.85cm] (ros40) {
                    \includegraphics[]{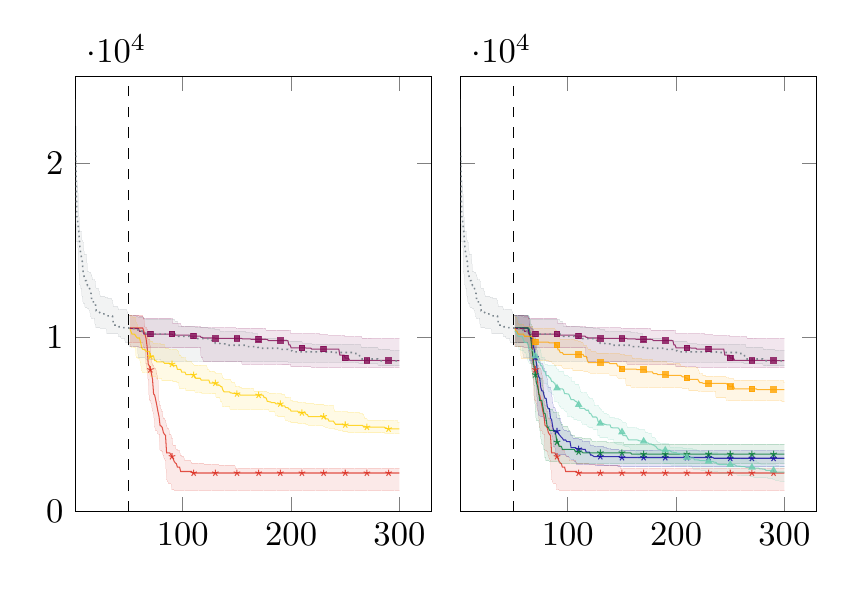}
                };

                \node[anchor=north west, yshift=-3cm, xshift=0.5cm] (label_rast20) {\textbf{Rastrigin 20D}};
                \node[anchor=west, yshift=-6cm] (rast20) {
                    \includegraphics[]{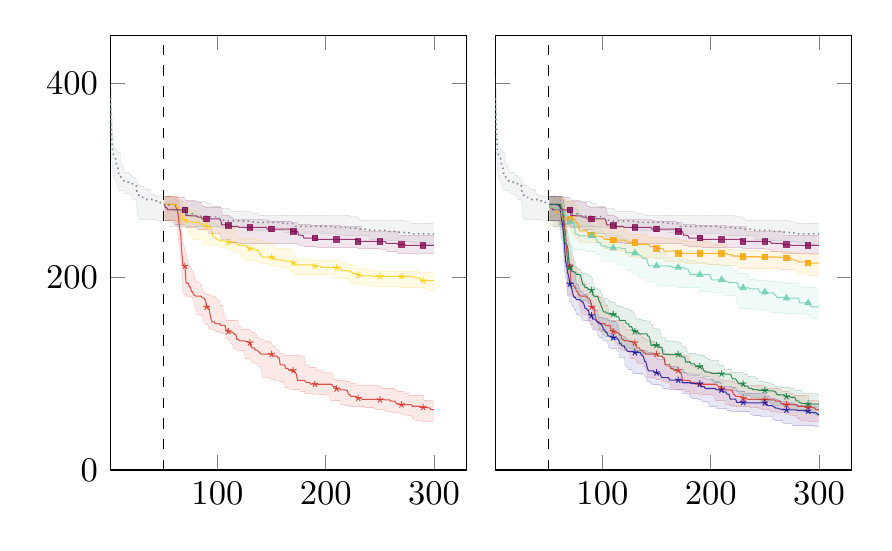}
                };

                \node[anchor=north west, yshift=-3cm, xshift=9.35cm] (label_rast40) {\textbf{Rastrigin 40D}};
                \node[anchor=west, yshift=-6cm, xshift=8.5cm] (rast40) {
                    \includegraphics[]{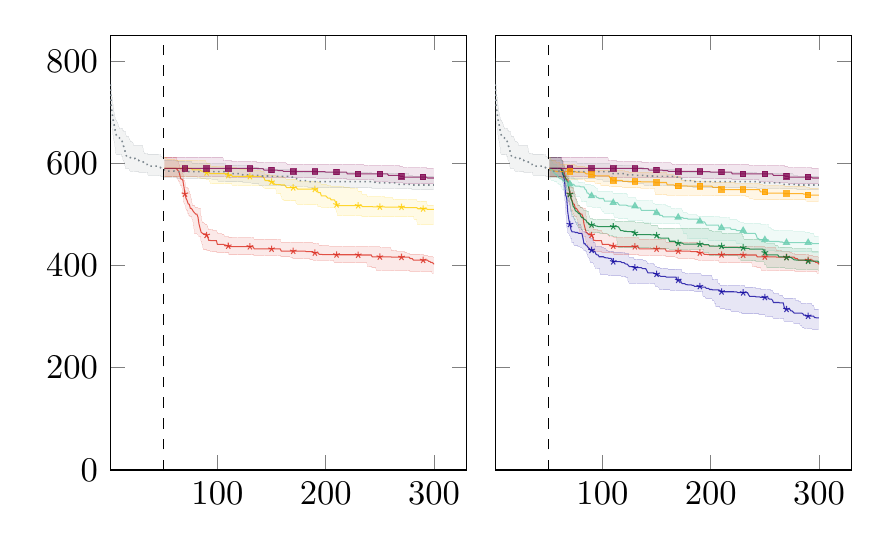}
                };

                \node[anchor=west, yshift=-9.6cm, xshift=0.2cm] (legend) {
                    \includegraphics[trim={0 0 0 5.5cm},clip,width=0.8\paperwidth]{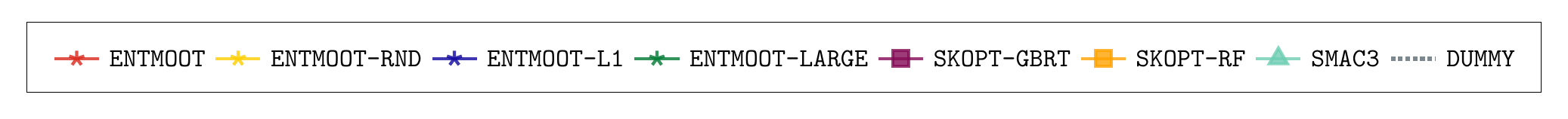}
                };

                % add axis labels
                \node[anchor=south, rotate=90, yshift=-0.5cm] at (ros20.west) (labelY) {\footnotesize Objective};

                \node[anchor=north, xshift=0.4cm, yshift=0.5cm] at (ros20.south) (labelX) {\footnotesize Number of evaluations};

                \node[anchor=north, xshift=0.4cm, yshift=0.5cm] at (ros40.south) (labelX) {\footnotesize Number of evaluations};

                \node[anchor=south, rotate=90, yshift=-0.5cm] at (rast20.west) (labelY) {\footnotesize Objective};

                \node[anchor=north, xshift=0.4cm, yshift=0.5cm] at (rast20.south) (labelX) {\footnotesize Number of evaluations};

                \node[anchor=north, xshift=0.4cm, yshift=0.5cm] at (rast40.south) (labelX) {\footnotesize Number of evaluations};
            \end{tikzpicture}
        \end{center}
        \caption{Comparison of \texttt{ENTMOOT} to other tree model-based black-box optimization frameworks (see:
        Section~\ref{sec:numerical_studies}) for the Rosenbrock and Rastrigin benchmark function. For all black-box
        functions the left graph emphasizes comparison of uncertainty measure and optimization strategy. The right
        graph compares \texttt{ENTMOOT} against other tree model-based BO tools. Dashed line demarcates initial design.}
        \label{fig:entmoot_tree_model_study}
    \end{figure*}
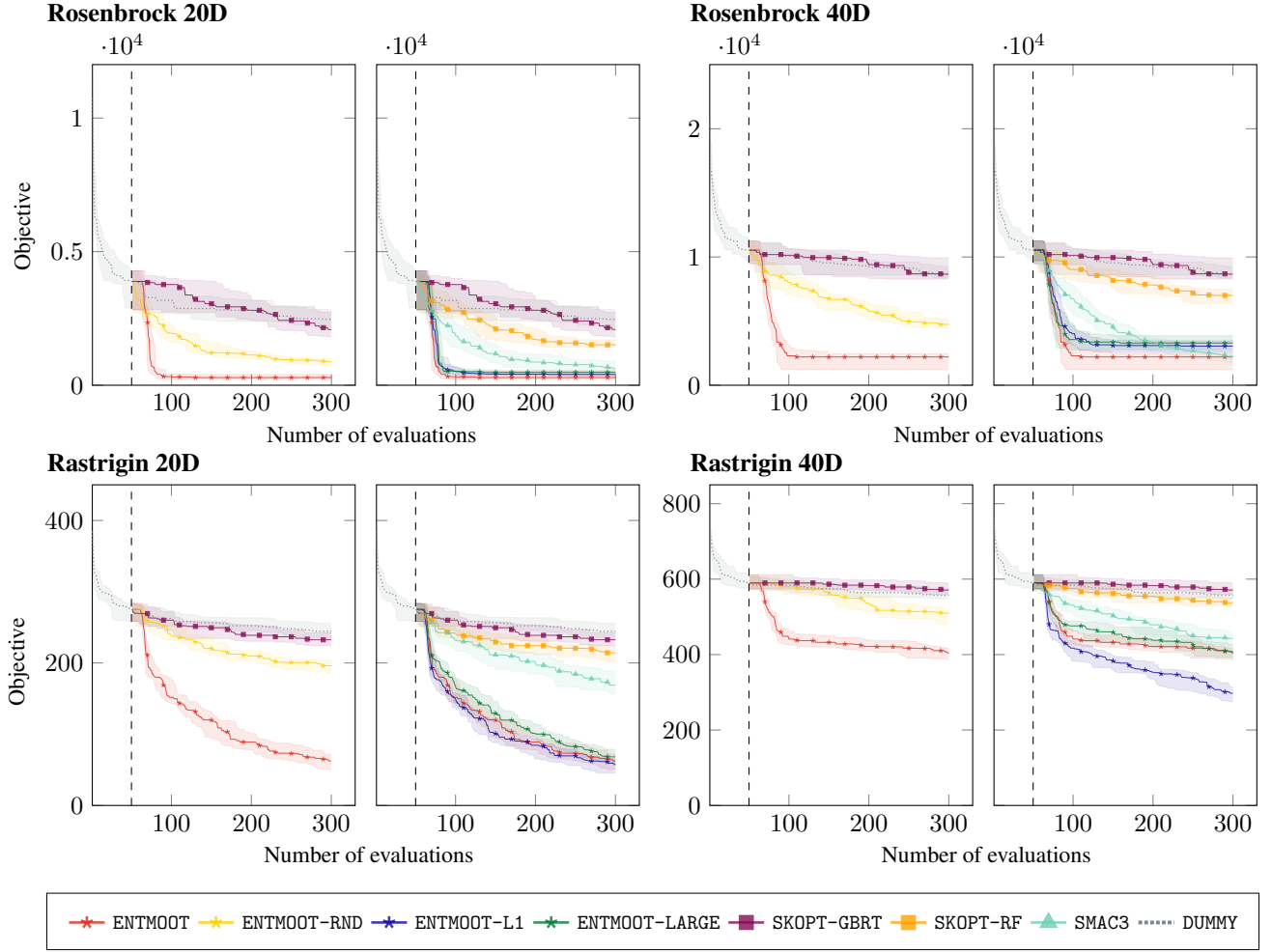

    \subsection{Comparison to other tree model-based frameworks} \label{sec:tree_study}
    This section tests two features that distinguish \texttt{ENTMOOT} from other tree model-based black-box
    optimization tools: (i) the distance-based measure to capture model uncertainty of tree-based models and (ii) the
    deterministic global optimization approach that avoids the need for stochastic methods to optimize the
    acquisition function. Fig.~\ref{fig:entmoot_tree_model_study} shows the per iteration best objective found by
    different algorithms on the 20D and 40D Rosenbrock and Rastrigin function. \texttt{ENTMOOT-RND} uses the same
    acquisition function as \texttt{ENTMOOT} but random sampling as an optimization strategy to determine new query
    points for black-box evaluation. This allows a better comparison to \texttt{SKOPT-GBRT} which uses the same
    underlying surrogate model and acquisition function optimization strategy, i.e.\ exactly the same random samples
    are used. Purely based on the exploration strategy, \texttt{ENTMOOT-RND} can find good solutions to the
    Rosenbrock function in both dimensional settings and constantly outperforms \texttt{SKOPT-GBRT}. Moreover, the
    default \texttt{ENTMOOT} setup, i.e.\ using deterministic global optimization to optimize the acquisition
    function, improves over both \texttt{SKOPT-GBRT} and \texttt{ENTMOOT-RND}. This emphasizes the importance of
    finding optimal trade-offs of exploitation and exploration when minimizing the acquisition function to determine
    new query points for black-box evaluations. All \texttt{ENTMOOT} variants perform similarly with
    \texttt{ENTMOOT-L1} providing the best objective value for 40D Rastrigin. For Rosenbrock 40D \texttt{SMAC3}
    manages to catch-up towards the end, outperforming both \texttt{ENTMOOT-LARGE} and \texttt{ENTMOOT-L1} but
    slightly staying behind the small tree model variant of \texttt{ENTMOOT}. Using other synthetic functions, e.g.\
    Sphere, lead to similar results and we refer to the appendices for more results as well as further
    details of the test setup.

    % performance study
    \begin{figure*}
        \begin{center}
            \begin{tikzpicture}
                \node[anchor=north west, yshift=3.2cm, xshift=0.5cm] (label_rover60) {\textbf{Rover 60D}};
                \node[anchor=west, xshift=0.08cm] (rover60) {
                    \includegraphics[]{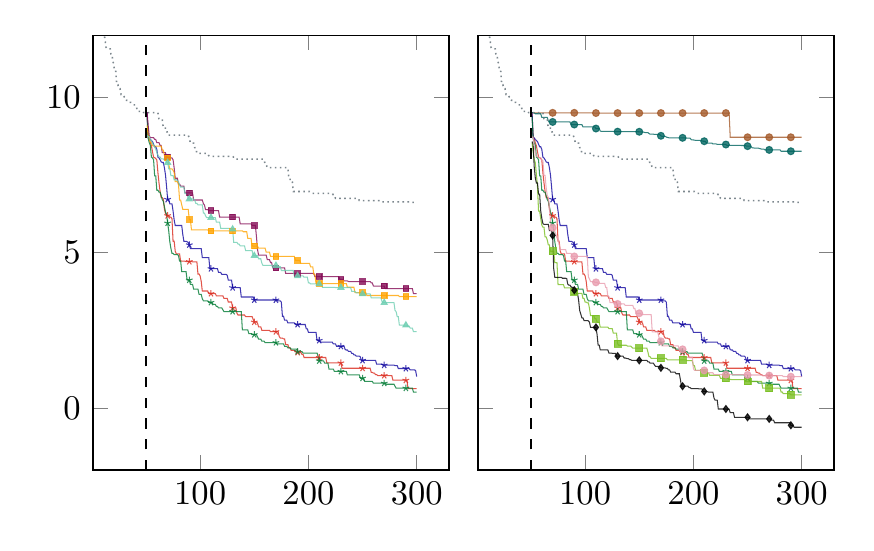}
                };

                \node[anchor=north west, yshift=3.2cm, xshift=9.35cm] (label_ferm80) {\textbf{Fermentation 80D}};
                \node[anchor=west, xshift=8.35cm] (ferm80) {
                    \includegraphics[]{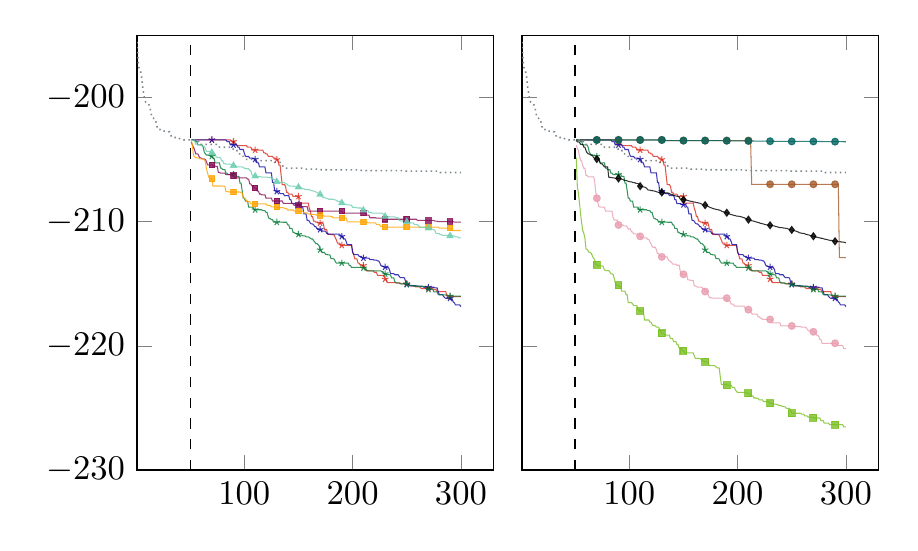}
                };

                \node[anchor=north west, yshift=-3cm, xshift=0.5cm] (label_ackley200) {\textbf{Ackley 200D}};
                \node[anchor=west, yshift=-6cm] (ackley200) {
                    \includegraphics[]{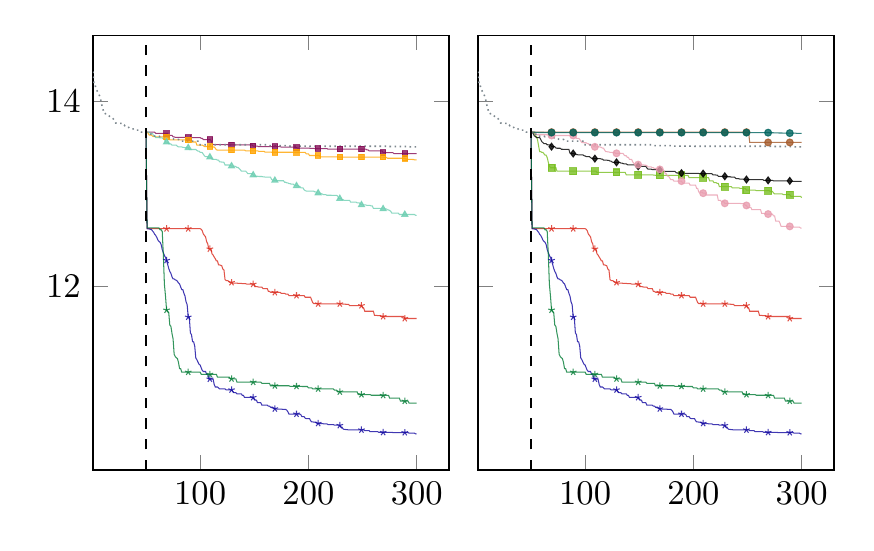}
                };

                \node[anchor=north west, yshift=-3cm, xshift=9.35cm] (label_robot14) {\textbf{Robot 14D}};
                \node[anchor=west, yshift=-6cm, xshift=8.70cm] (robot14) {
                    \includegraphics[]{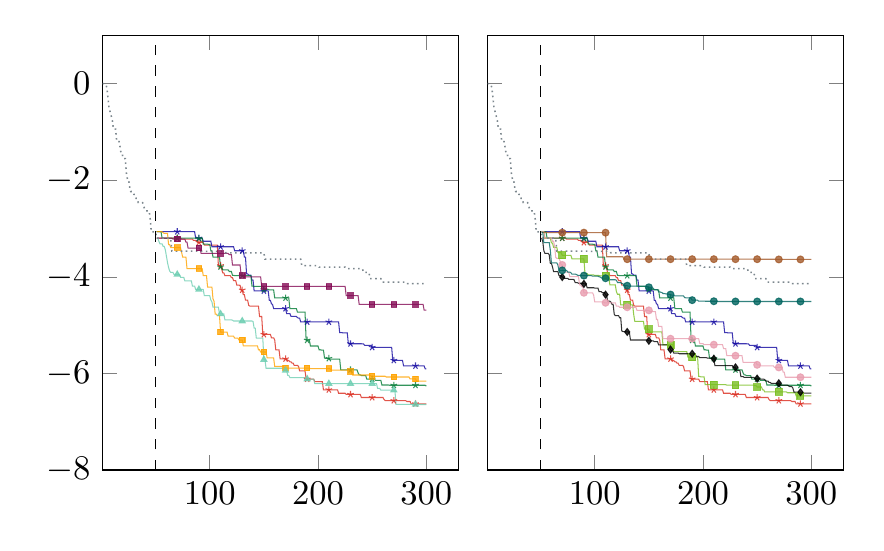}
                };

                \node[anchor=west, yshift=-9.9cm, xshift=1.2cm] (legend) {
                    \includegraphics[trim={0 0 0 5.5cm},clip,width=0.7\paperwidth]{figures/legend.pdf}
                };

                % add axis labels
                \node[anchor=south, rotate=90, yshift=-0.5cm] at (rover60.west) (labelY) {\footnotesize Objective};

                \node[anchor=north, xshift=0.4cm, yshift=0.5cm] at (rover60.south) (labelX) {\footnotesize Number of evaluations};

                \node[anchor=north, xshift=0.4cm, yshift=0.5cm] at (ferm80.south) (labelX) {\footnotesize Number of evaluations};

                \node[anchor=south, rotate=90, yshift=-0.5cm] at (ackley200.west) (labelY) {\footnotesize Objective};

                \node[anchor=north, xshift=0.4cm, yshift=0.5cm] at (ackley200.south) (labelX) {\footnotesize Number of evaluations};

                \node[anchor=north, xshift=0.4cm, yshift=0.5cm] at (robot14.south) (labelX) {\footnotesize Number of evaluations};
            \end{tikzpicture}
        \end{center}

        \caption{Comparison of \texttt{ENTMOOT} to other black-box optimization frameworks (see:
        Section~\ref{sec:numerical_studies}) using various benchmark functions. For all black-box functions the left
        graph compares \texttt{ENTMOOT} against other tree model-based BO tools. The right graph shows how
        \texttt{ENTMOOT} competes against other black-box optimization algorithms. Dashed line demarcates initial
        design. For graphs that also include confidence intervals we refer to Fig.~\ref{fig:entmoot_perf_study_conf_app}.}
        \label{fig:entmoot_performance_study}
    \end{figure*}

    \subsection{Comparison to other black-box optimization frameworks} \label{sec:perf_study}
    To compare \texttt{ENTMOOT}'s performance to other state-of-the-art black-box optimization solvers we test on
    various difficult instances. The figures compare \texttt{ENTMOOT} to other tree model-based approaches on the
    left-hand side and \texttt{ENTMOOT} against other black-box optimization frameworks on the right-hand side, i.e.\
    \texttt{SKOPT-GP}, \texttt{BOHAMIANN}, \texttt{CMA-ES}, \texttt{BFGS} and \texttt{NM}. We compare each
    algorithm's performance on the rover trajectory planning problem \citep{wang2018BatchedBO} which is a 60D
    benchmark instance. Fig.~\ref{fig:entmoot_performance_study} shows how all \texttt{ENTMOOT} variants outperform
    other tree-based model BO frameworks. In comparison to other black-box optimization frameworks, \texttt{ENTMOOT}
    is on par with \texttt{CMA-ES} and the Gaussian process based BO tool \texttt{SKOPT-GP}. \texttt{BOHAMIANN}
    outperforms \texttt{ENTMOOT} with a slightly better black-box objective at the end. \par
    Next, we compare on the fermentation model benchmark which is based on a mechanistic model
    \citep{znad2004KineticModel,elqotbi2013CFDModel}. This model describes the manufacturing $C_{P}$ of a chemical
    product based on a control sequence $\boldsymbol{x}$ consisting of 80 subsequent time steps for which the
    $k_{L}a$ value is determined. This value quantifies the efficiency of providing oxygen to the reaction and
    strongly influences the production. Integrating a system of four differential and one algebraic equation
    determines the production $C_{P}$ based on a control sequence $\boldsymbol{x}$. Here, the well-established
    \texttt{SKOPT-GP} algorithm based on Gaussian processes takes the lead followed by \texttt{CMA-ES}. However,
    \texttt{ENTMOOT} significantly outperforms other tree model-based approaches and the neural network approach
    \texttt{BOHAMIANN}. \par
    In the next study, we evaluate the 200D Ackley function on a domain $\left [ -5,10 \right ]^{200}$ with results
    given in Fig.~\ref{fig:entmoot_performance_study}. All \texttt{ENTMOOT} variants outperform other tree-based
    model BO frameworks after only a few iterations. \texttt{ENTMOOT-LARGE} and \texttt{ENTMOOT-L1} perform
    particularly well on this difficult benchmark problem. Also, in comparison with other black-box optimization
    tools, \texttt{ENTMOOT} performs best. The 200D Ackley function is notorious for exposing over-exploration in
    Gaussian process-based BO frameworks as observed by \citet{eriksson2019BOTR} which explains the poor performance
    of \texttt{SKOPT-GP}. \par
    Finally, we analyze a 14D robot pushing benchmark by \citet{wang2018BatchedBO} which is considered very
    challenging for black-box optimization frameworks due to high noise levels in its output and normally requires
    large evaluation budgets. Fig.~\ref{fig:entmoot_performance_study} shows how every algorithm struggles to
    consistently provide good solutions leading to large confidence intervals. When analyzing the median, some
    algorithms consistently perform worse than random sampling, e.g.\ \texttt{BFGS}, highlighting the difficulty of
    the benchmark problem. Comparing with other tree-based model frameworks, \texttt{SMAC3} comes out as the winner,
    providing a slightly better median solution than \texttt{ENTMOOT}. Compared to other approaches \texttt{ENTMOOT}
    performs slightly better than \texttt{BOHAMIANN} and \texttt{SKOPT-GP}. For more details on algorithms and
    benchmark functions, we refer to the appendices. We also provide more details regarding confidence
    intervals in Fig.~\ref{fig:entmoot_perf_study_conf_app}.

    \subsection{Large-scale extension} \label{sec:large_study}
    This section provides a proof of concept showing how better objective lower bounding strategies can help handle
    large-scale tree models. We use the concrete strength data set \cite{yeh1998ModelingNetworksc} from the UCI
    machine learning repository \cite{dua2017UCI}. The objective is to optimize concrete compressive strength based
    on ingredient proportions. We compare Gurobi~9 and \texttt{ENTMOOT} runs using advanced optimization strategies
    from Section~\ref{sec:decision_making_under_uncertainty}. Table~\ref{tab:concres} shows the bound improvements.
    \texttt{ENTMOOT} highly benefits from the warm-starting approach and bounding strategy mentioned in
    Section~~\ref{sec:decision_making_under_uncertainty} and produces better upper and lower bounds after 4~h of
    runtime. For blank entries in Table~\ref{tab:concres}, \texttt{ENTMOOT} has proven $\epsilon$-global optimality
    already, with a relative optimality gap of 0.01 \%. This happens for large penalty hyperparameter values, as
    large regions can be rejected quickly. In preliminary studies, we found better performance for
    Gurobi~9 when using the warm-starting strategy. However, for large instances this boost in
    performance is not sufficient to catch-up with our tailored lower-bounding strategy that exploits the
    tree structure of the underlying problem. This confirms the findings of \citet{mistry2018MixedIntegerEmbedded}.
    While Gurobi~9 is still the best choice for small and moderate instances,
    i.e.\ instances that commonly occur in design of experiments, we want to highlight that algorithmic elements
    exploiting the underlying structure of tree-based models can help when scaling the optimization efforts. Besides
    \citet{mistry2018MixedIntegerEmbedded} further improvements for optimizing tree-based models were proposed by
    \citet{misic2017OptimizationEnsembles}. For more details regarding the test setup and more results for different
    $\kappa$ values, see the appendices.

    \begin{table}[htb]
        \begin{center}
            \begin{tabular}{ |r||cc|cc||r||cc|cc| }
                \hhline{-----||-----}
                & \multicolumn{2}{c|}{ENTMOOT} & \multicolumn{2}{c||}{Gurobi~9.0} & & \multicolumn{2}{c|}{ENTMOOT}
                & \multicolumn{2}{c|}{Gurobi~9.0}
                \\
                \hhline{-----||-----}
                $\kappa=0.01$ & ub    & lb    & ub    & lb    & $\kappa=10$   & ub    & lb     & ub    & lb    \\
                \hhline{-----||-----}
                1 h           & -46.8 & -50.8 & -36.5 & -60.5 & 1 h           & -46.8 & -50.53 & -1.1  & -60.5 \\
                2 h           & -46.8 & -49.1 & -43.9 & -57.3 & 2 h           & -46.8 & -48.0  & -33.6 & -57.7 \\
                3 h           & -46.8 & -48.0 & -44.1 & -56.6 & 3 h           & -     & -      & -33.6 & -56.7 \\
                4 h           & -46.8 & -47.7 & -45.8 & -56.5 & 4 h           & -     & -      & -33.6 & -56.5 \\
                \hhline{=====::=====}
                $\kappa=0.1$  & ub    & lb    & ub    & lb    & $\kappa=100$  & ub    & lb     & ub    & lb    \\
                \hhline{-----||-----}
                1 h           & -46.8 & -50.8 & -40.9 & -60.5 & 1 h           & -46.8 & -49.7  & 143.9 & -60.5 \\
                2 h           & -46.8 & -49.1 & -43.9 & -57.4 & 2 h           & -     & -      & 143.9 & -57.3 \\
                3 h           & -46.8 & -47.8 & -45.0 & -56.6 & 3 h           & -     & -      & 0.5   & -56.5 \\
                4 h           & -46.8 & -47.6 & -45.0 & -51.9 & 4 h           & -     & -      & 0.5   & -51.8 \\
                \hhline{=====::=====}
                $\kappa=1$    & ub    & lb    & ub    & lb    & $\kappa=1000$ & ub    & lb     & ub    & lb    \\
                \hhline{-----||-----}
                1 h           & -46.8 & -50.8 & -40.9 & -60.5 & 1 h           & -46.8 & -49.7  & 143.9 & -60.5 \\
                2 h           & -46.8 & -49.1 & -43.9 & -57.4 & 2 h           & -     & -      & 143.9 & -57.3 \\
                3 h           & -46.8 & -47.8 & -45.0 & -56.6 & 3 h           & -     & -      & 0.5   & -56.5 \\
                4 h           & -46.8 & -47.6 & -45.0 & -51.9 & 4 h           & -     & -      & 0.5   & -51.8 \\
                \hhline{-----||-----}
            \end{tabular}
        \end{center}
        \caption{\label{tab:concres}Optimization results of \texttt{ENTMOOT} and Gurobi~9.0 for a GBT model that is
        trained on the concrete mixture design dataset. Blank entries refer to convergence of \texttt{ENTMOOT} with a
        relative optimality gap of 0.01 \%. Notation is: ub: upper objective bound, i.e.\ best feasible solution
        found, lb: lower objective bound, i.e.\ rigorous underestimator.}
    \end{table}

    \section{Conclusion}
    The success of tools like \texttt{SMAC3} and \texttt{SKOPT} shows that there is demand for using tree-based
    surrogate models in global black-box optimization tasks. \texttt{ENTMOOT} takes a new approach of handling
    tree-based models and tackles commonly-known drawbacks of tree-based models in black-box optimization
    \citep{shahriari2016BO}. By combining an intuitive distance-based uncertainty estimate with a deterministic
    global optimization strategy, we outperform other tree model-based approaches showing in detail how each of our
    proposals contributes to \texttt{ENTMOOT}'s good performance. On a comprehensive test on real-world complex
    tasks, \texttt{ENTMOOT} is shown to be a strong competitor to other black-box optimization frameworks, either
    outperforming them or providing competitive results. In most cases \texttt{ENTMOOT} finds excellent solutions
    after only a few iterations. Moreover, we show how algorithmic improvements can be used to scale the
    \texttt{ENTMOOT} approach to large data sets and tree models.\par
    For the future, we will show how \texttt{ENTMOOT}'s support of additional constraints that can be added to the
    optimization problem helps to include domain knowledge and safety critical requirements, potentially allowing
    better solutions given less observations. Such constraints can naturally be added to \texttt{ENTMOOT}'s
    deterministic optimization model encoding and solutions guarantee to satisfy the added constraints. Moreover, we
    hope that the work presented here motivates more efforts in exploring powerful tree-based models in Bayesian
    optimization settings and applications in chemical engineering.

    \section{Acknowledgements}
    The support of BASF SE, Ludwigshafen am Rhein, the EPSRC Centre for Doctoral Training in High Performance
    Embedded and Distributed Systems to M.M. (HiPEDS, EP/L016796/1), the Newton International Fellowship by the Royal
    Society (NIF\textbackslash R1\textbackslash 182194) to J.K., the grant by the Swedish Cultural Foundation in
    Finland to J.K. and an EPSRC Research Fellowship to R.M. (EP/P016871/1) is gratefully acknowledged.

    \clearpage

    \pagebreak

    \appendix

    \newcommand{\appFigScale}{0.98}

    \setlength\tabcolsep{2.3pt}
    \renewcommand{\arraystretch}{1.0}

    \section{List of Variables, Parameters and Sets}
    \begin{minipage}[a]{0.55\textwidth}
        \setstretch{1.0}
        \begin{tabularx}{\textwidth}{ |c|X| }
            \hline
            Symbol                      & Description
            \\
            \hline
            $f$                         & black-box function
            \\
            $\boldsymbol{x}$            & feature space
            \\
            $n$                         & number of features
            \\
            $\boldsymbol{x_{next}}$     & next proposed point to evaluate
            \\
            $\boldsymbol{x^*}$          & global minimizer of $f$
            \\
            $g(\boldsymbol{x})$         & equality constraints
            \\
            $h(\boldsymbol{x})$         & inequality constraints
            \\
            $\alpha$                    & uncertainty quantification
            \\
            $\kappa$                    & acquisition hyperparameter
            \\
            $\zeta$                     & exploration hyperparameter
            \\
            $\hat{\mu}$                 & mean of tree model
            \\
            $z_{t,l}$                   & variable of leaf $l$ in tree $t$
            \\
            $y_{i(s),j(s)}$             & binary variable of split $s$ in tree model
            \\
            $F_{t,l}$                   & value of leaf $l$ in tree $t$
            \\
            $\text{dist}$               & distance variable
            \\
            $\alpha_{limit}$            & parameter that limits $\alpha$
            \\
            $\epsilon$                  & optimality gap of MIP solver
            \\
            $v^L$                       & lower bounds of $x$
            \\
            $v^U$                       & upper bounds of $x$
            \\
            $v$                         & split interval thresholds
            \\
            $\sigma_{\mathscr{D},diag}$ & diagonal matrix containing standard deviation of data points in
            $\mathscr{D}$                              \\
            $\mu_{\mathscr{D}}$         & vector containing mean of data points in $\mathscr{D}$
            \\
            $y_{\mathscr{D}}$           & target values in dataset $\mathscr{D}$
            \\
            $r^{d,+}$                   & auxiliary variables taking positive contributions of Manhattan distance for
            data point $d \in \mathscr{D}$ \\
            $r^{d,-}$                   & auxiliary variables taking negative contributions of Manhattan distance for
            data point $d \in \mathscr{D}$ \\
            $M$                         & sufficiently large value to allow switching constraints
            \\
            $b$                         & binary variable switch between data points or cluster centers
            \\
            $S$                         & domain $S$ defined by node of branch-and-bound tree
            \\
            $\hat{R}^S$                 & lower bound for domain $S$
            \\
            $b^{\hat{\mu},S}$           & lower bound for tree model in domain $S$
            \\
            $b^{\hat{\alpha},S}$        & lower bound for tree model uncertainty in domain $S$
            \\
            $\boldsymbol{x_{feas}}$     & initial feasible solution
            \\
            $C_P$                       & production of chemical product for fermentation benchmark
            \\
            $k_La$                      & quantifies oxygen delivery efficiency for fermentation benchmark
            \\
            \hline
        \end{tabularx}
    \end{minipage}
    \hfill
    \begin{minipage}[b]{0.4\textwidth}
        \setstretch{1.0}
        \begin{tabularx}{\textwidth}{ |c|X| }
            \hline
            Index & Description                               \\
            \hline
            $d$   & data point $d \in \mathscr{D}$            \\
            $t$   & trees $t \in \mathscr{T}$                 \\
            $l$   & leaves $l \in \mathscr{L}_t$              \\
            $s$   & splits $s \in \mathscr{V}_t$              \\
            $i$   & vector element $i \in \left[ n \right]$   \\
            $j$   & vector element $j \in \left[ m_i \right]$ \\
            $k$   & cluster $k \in \mathscr{K}$               \\
            \hline
        \end{tabularx}

        \vspace{1.2cm}

        \begin{tabularx}{\textwidth}{ |c|X| }
            \hline
            Set                  & Description                                               \\
            \hline
            $\mathscr{D}$        & dataset                                                   \\
            $p$                  & order of norm $\lVert \cdot \rVert_{p}$                   \\
            $\Omega$             & constraints defined by tree model and linking constraints \\
            $\mathscr{T}$        & index set for trees of tree ensemble                      \\
            $\mathscr{L}_t$      & index set for leaves of tree $t$                          \\
            $\mathscr{V}_t$      & index set for splits of tree $t$                          \\
            $\text{Left}_{t,s}$  & index set of leaves left to split $s$ in tree $t$         \\
            $\text{Right}_{t,s}$ & index set of leaves right to split $s$ in tree $t$        \\
            $\mathscr{K}$        & set of clusters                                           \\
            \hline
        \end{tabularx}
    \end{minipage}

    \section{ENTMOOT details} \label{app:entmoot_details}
    In our numerical studies, we use various \texttt{ENTMOOT} settings to show how different features contribute to
    the performance of the framework. \texttt{ENTMOOT} uses LightGBM \citep{ke2017LightGBM} to train gradient-boosted
    tree models. The standard \texttt{ENTMOOT} variant as well as \texttt{ENTMOOT-L1} and \texttt{ENTMOOT-RND} use
    400 decision trees per ensemble with an interaction depth of 3. \texttt{ENTMOOT-LARGE} uses an ensemble of 800
    decision trees with an interaction depth of 2. Moreover, the minimum number of samples in one leaf is set to 20
    and the maximum number of leaves per tree is fixed at 5. All other hyperparameters of LightGBM are left at
    default values. \texttt{ENTMOOT-L1} uses the Manhattan distance as an uncertainty estimate which is formally
    introduced in Section~3.2.3. All other \texttt{ENTMOOT} variants use the squared Euclidean distance uncertainty
    measure defined in Section~3.2.2. The uncertainty term in the acquisition function is weighted with $\kappa$
    which is set to a value of 1.96 for all tests, as this is the default value of $\kappa$ for the \texttt{SKOPT}
    LCB acquisition function. The uncertainty bound hyperparameter $\zeta$ is set to 0.5 for all tests.
    \texttt{ENTMOOT-RND} uses the same random sampling optimization strategy as \texttt{SKOPT} to minimize the
    acquisition function. In fact, due to the fixed random seeds in every run, \texttt{ENTMOOT-RND} uses exactly the
    same sampling points to minimize the acquisition function as \texttt{SKOPT-GBRT} and \texttt{SKOPT-RF}. This
    allows a direct comparison of the models implemented in both frameworks. All other \texttt{ENTMOOT} variations
    encode the tree ensemble and the uncertainty estimate as a mathematical model by formulating a MINLP. The model
    is solved using Gurobi~9 \citep{gurobi}, a commercial optimization software. Gurobi~9 is set to a time limit of 2
    minutes with most runs finishing at the default relative optimality gap of 0.01 \%.

    \section{Other algorithms}
    This section describes details regarding other algorithms that we compare against \texttt{ENTMOOT} and what
    specifications were used to run numerical studies. \texttt{DUMMY} randomly samples the search space to find
    optimal points of benchmark functions. The first 50 points of \texttt{DUMMY} initialize all other algorithms. We
    define a global random state for every run and use it as the random seed for \texttt{DUMMY} and all
    non-deterministic algorithms that require a fixed seed for reproducible results. \par

    \subsection{\texttt{BOHAMIANN}}
    The \texttt{BOHAMIANN} framework uses Bayesian neural networks and was first introduced by
    \citet{springenberg2016Bohamiann}. We use the \texttt{RoBo} v0.2.0 implementation available at:
    \url{https://automl.github.io/RoBO/}. All hyperparameters of \texttt{BOHAMIANN} are left at default values. \par

    \subsection{\texttt{CMA-ES}}
    For comparisons against \texttt{CMA-ES} \citep{hansen2006CMA} we use the \texttt{cma} v3.0.3 implementation
    available at: \url{https://pypi.org/project/cma/}. As an initial value $\boldsymbol{x_{0}}$, we provided the best
    point obtained from the initial 50 random values provided by \texttt{DUMMY}. To allow the best possible
    performance of the framework, we follow recommendations of the user manual and wrap the black-box function to
    allow standardized inputs. We use the object \textit{CMAEvolutionStrategy} and, based on the scaling, set
    \texttt{sigma0} to 0.5.

    \subsection{\texttt{SMAC3}}
    We used the Python implementation \texttt{SMAC} v3 (0.13) \citep{hutter2011SequentialModel} by the AutoML group
    which is available at: \url{https://github.com/automl/SMAC3}. To get comparable results, we used the
    \texttt{SMAC4HPO} object which uses random forests as the underlying surrogate model. Given the examples that are
    tested, we set $\texttt{run\_obj=``quality''}$ and $\texttt{deterministic=``true''}$ as suggested in the user
    manual. All other hyperparameters are left at default values.

    \subsection{\texttt{SKOPT}}
    \texttt{Scikit-Optimize} (\texttt{SKOPT}) \citep{skopt2018} is a popular BO library which supports various
    surrogate models. The numerical studies use the most recent implementation \texttt{SKOPT} v0.8 which is available
    at: \url{https://github.com/scikit-optimize/scikit-optimize}. \texttt{SKOPT-GBRT} and \texttt{SKOPT-RF} refer to
    the \texttt{Scikit-Optimize} variants that use GBRTs and RFs as surrogate models, respectively. \texttt{SKOPT-GP}
    uses GPs as the underlying machine learning model. Besides specifying the surrogate model, all tests were
    performed with default hyperparameter settings.\par

    \subsection{\texttt{NELDER-MEAD}, \texttt{BFGS}}
    Both algorithms are implemented as functions in \texttt{SciPy} v1.5.3 \citep{jones2001Scipy} which we used for
    testing. We lowered the solution tolerances to prevent premature termination, so that both algorithms can make
    full use of the black-box evaluation budget. For $\boldsymbol{x_{0}}$, we provide the best point obtained from
    the initial 50 random values.

    \section{Numerical studies}

    \subsection{Benchmark functions}
    This section provides more information for all benchmark functions used. For more detailed information, we refer
    to the original papers.

    \subsubsection{Synthetic functions}
    We use the following synthetic functions \citep{surjanovic2020Func} for testing: Rosenbrock, Rastrigin, Sphere
    and Styblinski-Tang. We evaluate Rosenbrock on the domain $\left[-2.048,2.048\right]^{dim}$, Rastrigin on
    $\left[-5.12,5.12\right]^{dim}$, Sphere on $\left[-5.12,5.12\right]^{dim}$ and Styblinski-Tang on $\left[-5.00,5
    .00\right]^{dim}$. For these functions, we test different dimensionality settings $dim \in \{10,20,40\}$. We test
    Ackley 200D on the domain $\left[-5.00,10.00\right]^{200}$.

    \subsubsection{Rover trajectory planning} \label{sec:rover}
    We use a 60D rover trajectory planning problem first proposed by \citet{wang2018BatchedBO}. The challenge is to
    fit 30 points on a 2D plane to a B-spline that defines the rover trajectory. The reward is computed based on
    which terrain the trajectory intersects according to the function: $f(\boldsymbol{x})= c(\boldsymbol{x}) - 10
    (\lVert \boldsymbol{x}_{1,2} - \boldsymbol{x}_{s} \rVert_{1} + \lVert \boldsymbol{x}_{59,60}-\boldsymbol{x}_{g}
    \rVert_{1} ) + 5$. More specifically, every collision with rough terrain adds -20 to $c(\boldsymbol{x})$ as a
    penalty. $\boldsymbol{x}_{s}$ and $\boldsymbol{x}_{g}$ are predefined start and end position and a penalty is
    added if these are not matched. Instead of maximizing the reward, we minimize the negative reward.

    \subsubsection{Robot pushing}
    A robot pushing benchmark is used from \citet{wang2018BatchedBO}. A controller is tuned for two robot hands to
    move objects to a specified target location. The search space is 14D and the following hyperparameters are tuned:
    location and rotation of hands, speed of movement, direction of movement and pushing time. The reward function
    is: $f(\boldsymbol{x}) = \sum\limits_{i \in \{1,2\}} \Vert \boldsymbol{x}_{gi} - \boldsymbol{x}_{si} \rVert -
    \lVert \boldsymbol{x}_{gi} - \boldsymbol{x}_{fi} \rVert$. $\boldsymbol{x}_{si}$, $\boldsymbol{x}_{fi}$ and
    $\boldsymbol{x}_{gi}$ define initial, final and goal location of objects, respectively. Instead of maximizing the
    reward we minimize the negative reward.

    \subsection{Data distance as an effective measure for uncertainty} \label{app:d2}
    The Section~\ref{sec:uncertainty_study} study shows \texttt{ENTMOOT}'s uncertainty estimate at work. The purpose
    is to analyze how well the distance-based concepts defined in
    Section~\ref{sec:encoding_euclidean_distance_squared_uncertainty} and
    Section~\ref{sec:encoding_manhattan_distance_uncertainty} can capture uncertainty in tree model predictions.
    For analysis, we can change the value of $\kappa$ and thereby force the optimal solution to be closer or further
    away from training data. \par
    For this trial, the Gurobi~9 time limit is set to 1 h, with most runs finishing beforehand at the default
    relative optimality gap of 0.01 \%. A set $\kappa \in \left\{0.5,1.0,\dots,10.0\right\}$ is evaluated and 30
    individual runs per method with random states $rnd \in \left\{101,102,103,\dots,130\right\}$ define the
    confidence intervals. We use an initial data set size of 200 randomly generated samples and compute
    $\boldsymbol{x}_{next}$, i.e.\ the next black-box evaluation proposal $f(\boldsymbol{x}_{next})$ by
    \texttt{ENTMOOT} with the modified mathematical model. We calculate the relative model error according to:
    % EQUATION: relative error
    \begin{equation}
        \label{eq:averageerror}
        \epsilon_{\hat{\mu}}(\kappa) = \left\lVert \frac{\hat{\mu}(\boldsymbol{x}_{next})-f(\boldsymbol{x}_{next})
        }{\hat{\mu}(\boldsymbol{x}_{next})}\right\rVert,
    \end{equation}
    with $\hat{\mu}$ defining the tree model prediction. The relative model error $\epsilon_{\hat{\mu}}$ decreases
    with growing values for $\kappa$. Eventually, $\kappa$ becomes too large and forces $\boldsymbol{x}_{next}$ to
    take the location of the best available data point. In Fig.~\ref{fig:entmoot_uncertainty_study}, this behavior is
    shown at end of the graph when the relative model error converges asymptotically. With a higher penalty, $\alpha$
    dominates the objective in Equ.~\eqref{eq:objective_alt} and tree model predictions at $\boldsymbol{x}_{next}$
    become larger and more conservative. Empirically, this shows that the proposed uncertainty estimate works and
    both measures, introduced in Section~\ref{sec:encoding_euclidean_distance_squared_uncertainty} and
    Section~\ref{sec:encoding_manhattan_distance_uncertainty}, are good indicators for model uncertainty of
    tree-based models.

    \subsection{Comparison to other tree-based model frameworks} \label{app:d3}
\begin{figure*}
        \begin{center}
            \begin{tikzpicture}
                \node[anchor=north west, yshift=3.2cm, xshift=0.4cm] (label_ros10) {\textbf{Rosenbrock 10D}};
                \node[anchor=west, xshift=0cm] (ros10) {
                    \includegraphics[scale=\appFigScale]{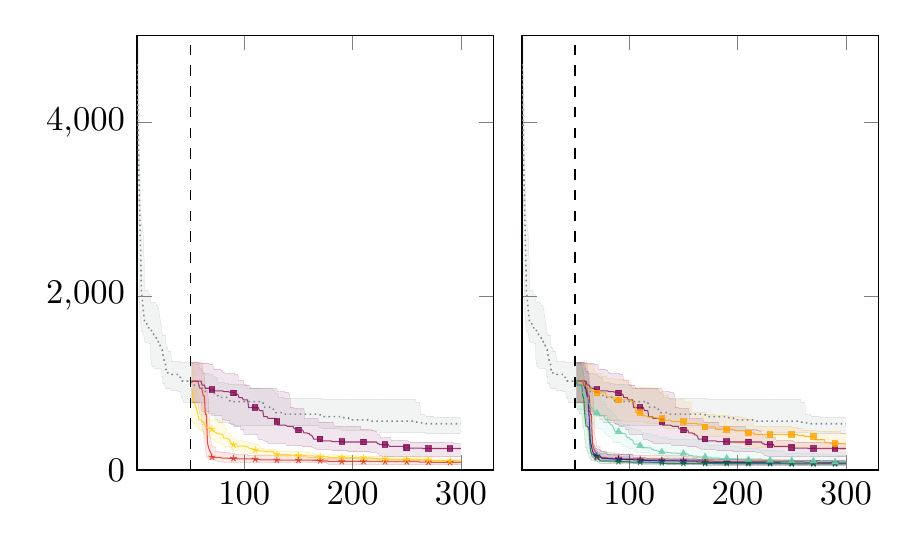}
                };

                \node[anchor=north west, yshift=-3cm, xshift=0.4cm] (label_ros20) {\textbf{Rosenbrock 20D}};
                \node[anchor=west, yshift=-6.2cm, xshift=0.35cm] (ros20) {
                    \includegraphics[scale=\appFigScale]{figures/tree_ros_20.pdf}
                };

                \node[anchor=north west, yshift=-9.2cm, xshift=0.4cm] (label_ros40) {\textbf{Rosenbrock 40D}};
                \node[anchor=west, yshift=-12.4cm, xshift=0.625cm] (ros40) {
                    \includegraphics[scale=\appFigScale]{figures/tree_ros_40.pdf}
                };

                \node[anchor=north west, yshift=3.2cm, xshift=9.5cm] (label_ras10) {\textbf{Rastrigin 10D}};
                \node[anchor=west, xshift=8.6cm] (ras10) {
                    \includegraphics[scale=\appFigScale]{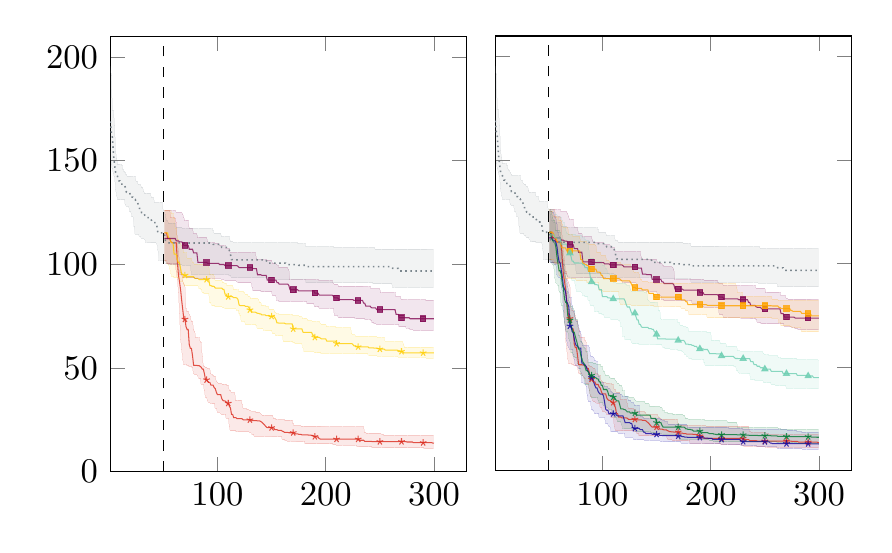}
                };

                \node[anchor=north west, yshift=-3cm, xshift=9.35cm] (label_ras20) {\textbf{Rastrigin 20D}};
                \node[anchor=west, yshift=-6.42cm, xshift=8.6cm] (ras20) {
                    \includegraphics[scale=\appFigScale]{figures/tree_rast_20.pdf}
                };

                \node[anchor=north west, yshift=-9.2cm, xshift=9.35cm] (label_ras40) {\textbf{Rastrigin 40D}};
                \node[anchor=west, yshift=-12.62cm, xshift=8.6cm] (ras40) {
                    \includegraphics[scale=\appFigScale]{figures/tree_rast_40.pdf}
                };

                \node[anchor=west, yshift=-16.0cm, xshift=0.2cm] (legend) {
                    \includegraphics[trim={0 0 0 5.5cm},clip,width=0.8\paperwidth]{figures/legend.pdf}
                };

                % add axis labels
                \node[anchor=south, rotate=90, yshift=-0.5cm] at (ros10.west) (labelY) {\footnotesize Objective};

                \node[anchor=north, xshift=0.4cm, yshift=0.5cm] at (ros10.south) (labelX) {\footnotesize Number of evaluations};

                \node[anchor=south, rotate=90, yshift=-0.5cm] at (ros20.west) (labelY) {\footnotesize Objective};

                \node[anchor=north, xshift=0.4cm, yshift=0.5cm] at (ros20.south) (labelX) {\footnotesize Number of evaluations};

                \node[anchor=south, rotate=90, yshift=-0.5cm] at (ros40.west) (labelY) {\footnotesize Objective};

                \node[anchor=north, xshift=0.4cm, yshift=0.5cm] at (ros40.south) (labelX) {\footnotesize Number of evaluations};

                \node[anchor=north, xshift=0.4cm, yshift=0.5cm] at (ras10.south) (labelX) {\footnotesize Number of evaluations};

                \node[anchor=north, xshift=0.4cm, yshift=0.5cm] at (ras20.south) (labelX) {\footnotesize Number of evaluations};

                \node[anchor=north, xshift=0.4cm, yshift=0.5cm] at (ras40.south) (labelX) {\footnotesize Number of evaluations};
            \end{tikzpicture}
        \end{center}
        \caption{Comparison of \texttt{ENTMOOT} to other tree model-based black-box optimization frameworks (see:
        Section~\ref{sec:numerical_studies}) for the Rosenbrock and Rastrigin benchmark functions. For all black-box
        functions the left graph emphasizes comparison of uncertainty measure and optimization strategy. The right
        graph compares \texttt{ENTMOOT} against other tree model-based BO tools. Dashed line demarcates initial design.}
        \label{fig:entmoot_tree_model_study_app_1}
    \end{figure*}

\begin{figure*}
        \begin{center}
            \begin{tikzpicture}
                \node[anchor=north west, yshift=3.2cm, xshift=0.4cm] (label_sphere10) {\textbf{Sphere 10D}};
                \node[anchor=west, xshift=0.01cm] (sphere10) {
                    \includegraphics[scale=\appFigScale]{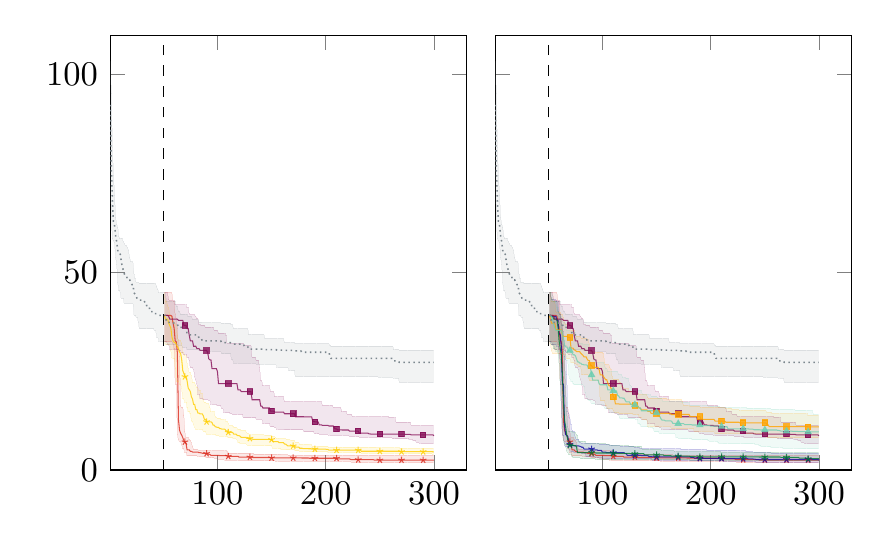}
                };

                \node[anchor=north west, yshift=-3cm, xshift=0.4cm] (label_sphere20) {\textbf{Sphere 20D}};
                \node[anchor=west, yshift=-6.2cm, xshift=0.01cm] (sphere20) {
                    \includegraphics[scale=\appFigScale]{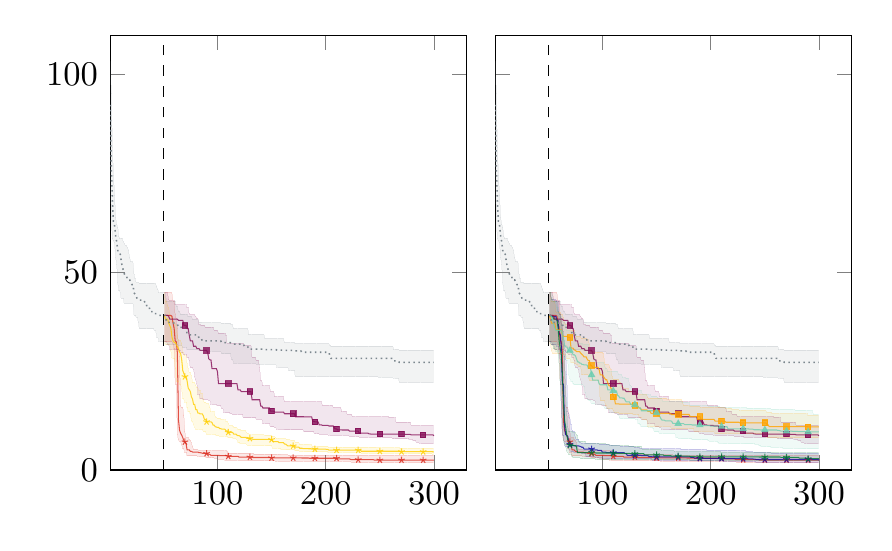}
                };

                \node[anchor=north west, yshift=-9.2cm, xshift=0.4cm] (label_sphere40) {\textbf{Sphere 40D}};
                \node[anchor=west, yshift=-12.4cm, xshift=0.01cm] (sphere40) {
                    \includegraphics[scale=\appFigScale]{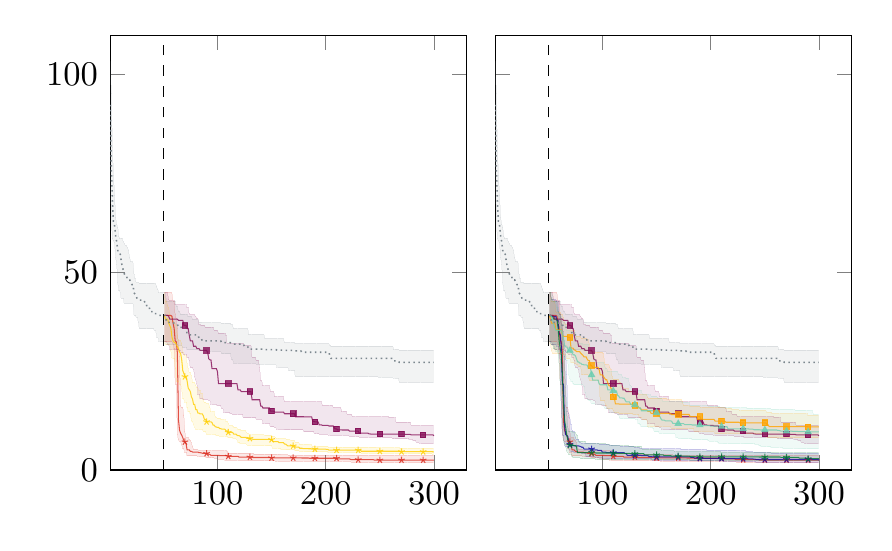}
                };

                \node[anchor=north west, yshift=3.2cm, xshift=9.5cm] (label_stang10) {\textbf{Styblinski-Tang 10D}};
                \node[anchor=west, xshift=8.37cm] (stang10) {
                    \includegraphics[scale=\appFigScale]{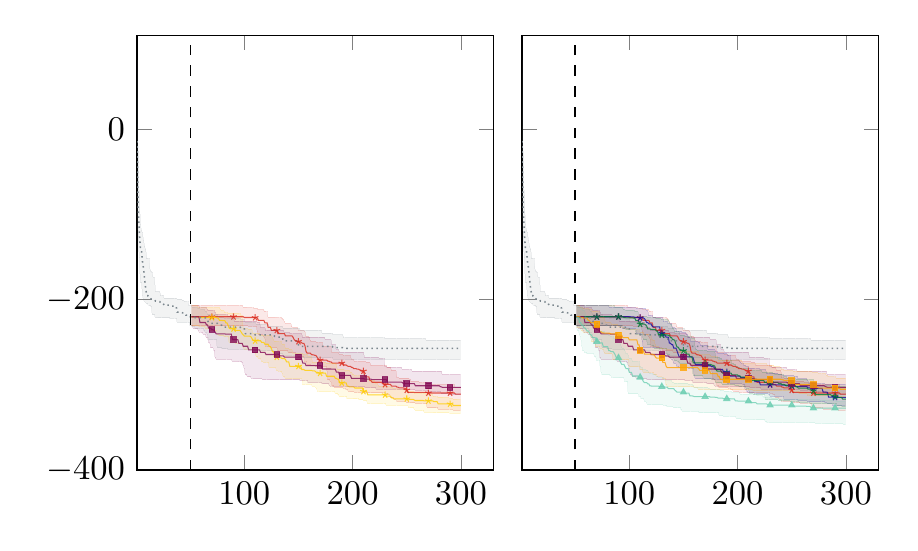}
                };

                \node[anchor=north west, yshift=-3cm, xshift=9.35cm] (label_stang20) {\textbf{Styblinski-Tang 20D}};
                \node[anchor=west, yshift=-6.2cm, xshift=8.37cm] (stang20) {
                    \includegraphics[scale=\appFigScale]{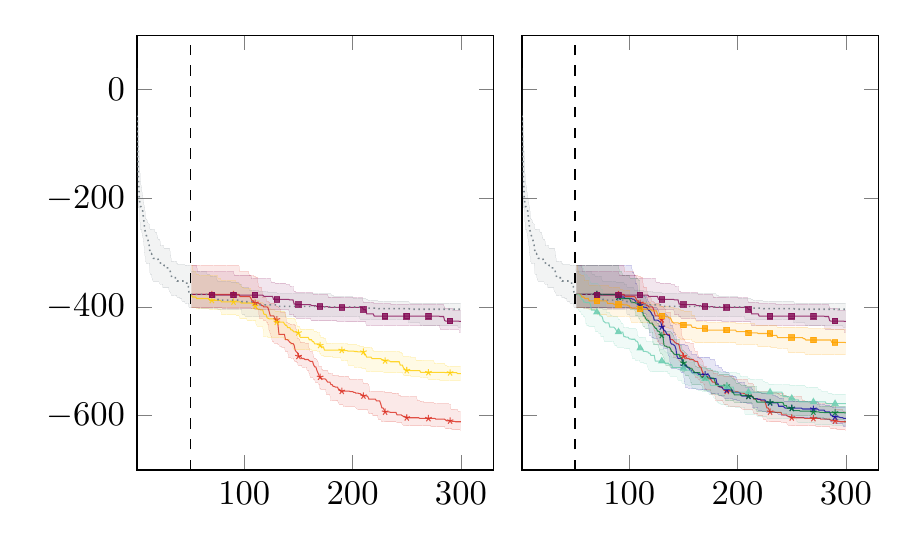}
                };

                \node[anchor=north west, yshift=-9.2cm, xshift=9.35cm] (label_stang40) {\textbf{Styblinski-Tang 40D}};
                \node[anchor=west, yshift=-12.4cm, xshift=8.1cm] (stang40) {
                    \includegraphics[scale=\appFigScale]{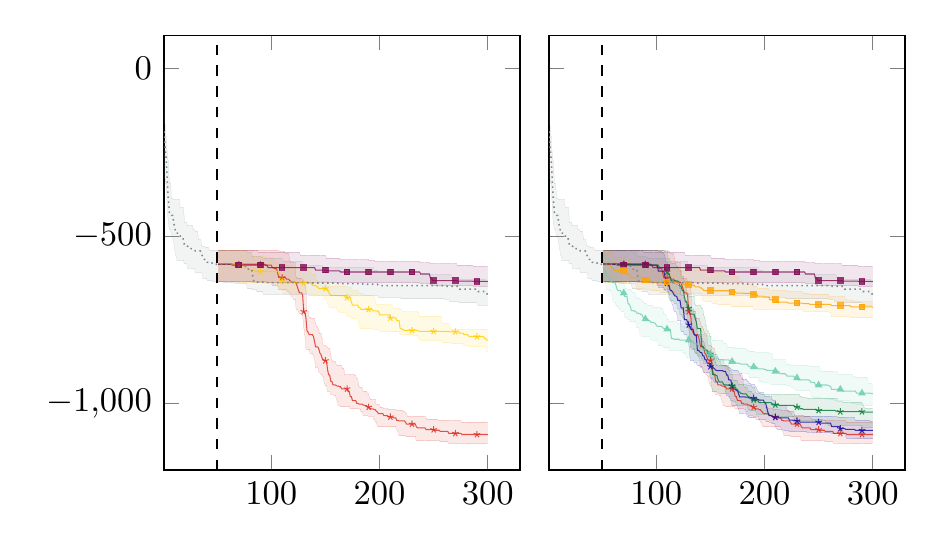}
                };

                \node[anchor=west, yshift=-16.0cm, xshift=0.2cm] (legend) {
                    \includegraphics[trim={0 0 0 5.5cm},clip,width=0.8\paperwidth]{figures/legend.pdf}
                };

                % add axis labels
                \node[anchor=south, rotate=90, yshift=-0.5cm] at (sphere10.west) (labelY) {\footnotesize Objective};

                \node[anchor=north, xshift=0.4cm, yshift=0.5cm] at (sphere10.south) (labelX) {\footnotesize Number of evaluations};

                \node[anchor=south, rotate=90, yshift=-0.5cm] at (sphere20.west) (labelY) {\footnotesize Objective};

                \node[anchor=north, xshift=0.4cm, yshift=0.5cm] at (sphere20.south) (labelX) {\footnotesize Number of evaluations};

                \node[anchor=south, rotate=90, yshift=-0.5cm] at (sphere40.west) (labelY) {\footnotesize Objective};

                \node[anchor=north, xshift=0.4cm, yshift=0.5cm] at (sphere40.south) (labelX) {\footnotesize Number of evaluations};

                \node[anchor=north, xshift=0.4cm, yshift=0.5cm] at (stang10.south) (labelX) {\footnotesize Number of evaluations};

                \node[anchor=north, xshift=0.4cm, yshift=0.5cm] at (stang20.south) (labelX) {\footnotesize Number of evaluations};

                \node[anchor=north, xshift=0.4cm, yshift=0.5cm] at (stang40.south) (labelX) {\footnotesize Number of evaluations};

            \end{tikzpicture}
        \end{center}
        \caption{Comparison of \texttt{ENTMOOT} to other tree model-based black-box optimization frameworks (see:
        Section~\ref{sec:numerical_studies}) for the Sphere and Styblinski-Tang benchmark functions. For all
        black-box functions the left graph emphasizes comparison of uncertainty measure and optimization strategy.
        The right graph compares \texttt{ENTMOOT} against other tree model-based BO tools. Dashed line demarcates
        initial design.}
        \label{fig:entmoot_tree_model_study_app_2}
    \end{figure*}

    This section presents additional results when comparing \texttt{ENTMOOT}'s performance to other tree-based model
    black-box optimization frameworks (see: Section~\ref{sec:tree_study}) for Rosenbrock, Rastrigin, Sphere and
    Styblinski-Tang benchmark functions given in Fig.~\ref{fig:entmoot_tree_model_study_app_1} and Fig
    .\ref{fig:entmoot_tree_model_study_app_2}. Similar to results presented in Section~\ref{sec:tree_study},
    \texttt{ENTMOOT} performs well against other algorithms, especially in high dimensional settings. To compute
    confidence intervals, we use random states $rnd \in \left\{101,102,103,\dots,130\right\}$.

    \subsection{Comparison to other black-box optimization frameworks} \label{app:d4}
    Confidence intervals in the Section~\ref{sec:perf_study} performance comparison of \texttt{ENTMOOT} with other
    black-box optimization tools use random states $rnd \in \left\{101,102,103,\dots,130\right\}$.

    \begin{figure*}
        \begin{center}
            \begin{tikzpicture}
                \node[anchor=north west, yshift=3.2cm, xshift=0.5cm] (label_rover60) {\textbf{Rover 60D}};
                \node[anchor=west, xshift=0.08cm, yshift=0.5cm] (rover60) {
                    \includegraphics[scale=0.90]{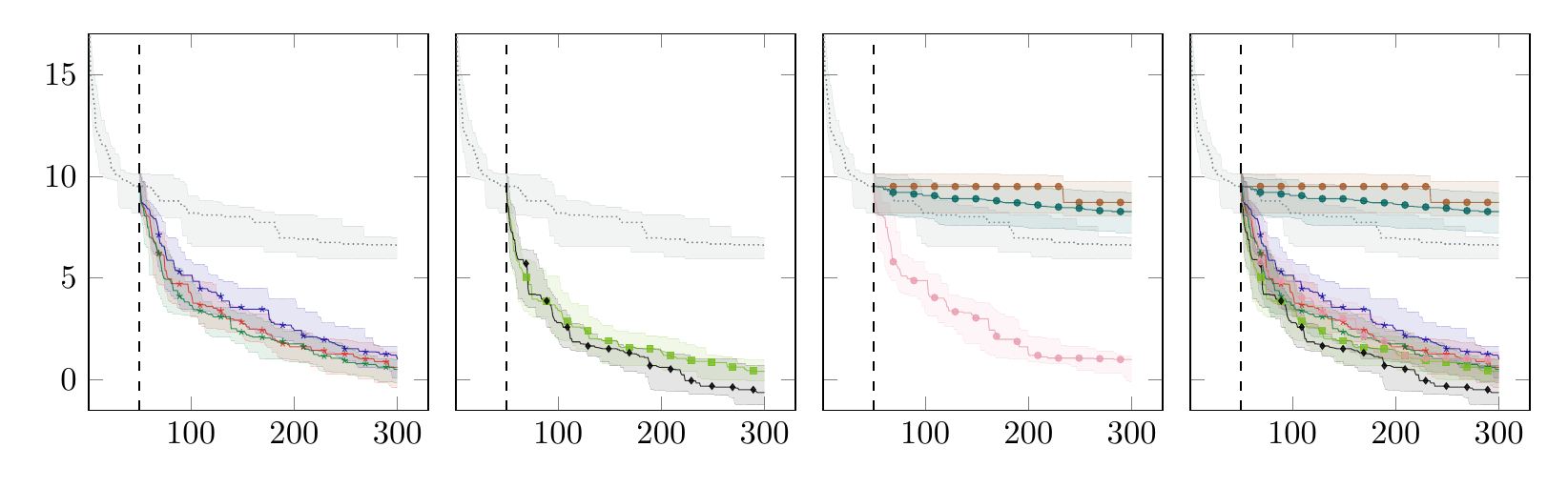}
                };

                \node[anchor=north west, yshift=-1.7cm, xshift=0.5cm] (label_ferm80) {\textbf{Fermentation 80D}};
                \node[anchor=west, yshift=-4.5cm, xshift=-0.32cm] (ferm80) {
                    \includegraphics[scale=0.90]{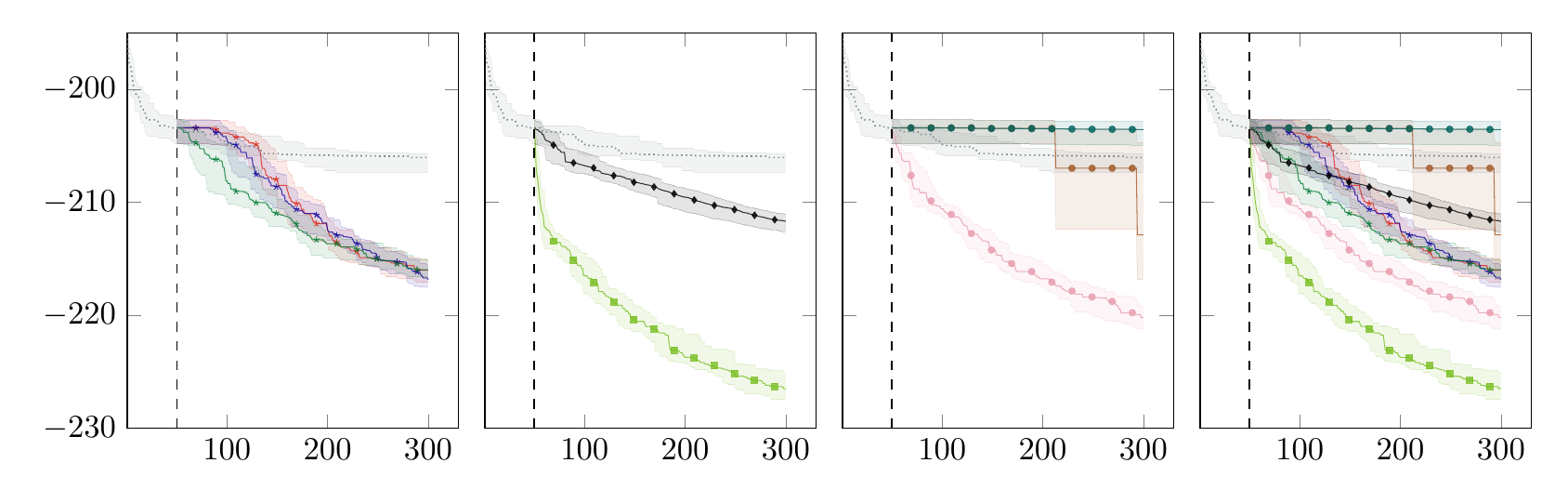}
                };

                \node[anchor=north west, yshift=-6.8cm, xshift=0.5cm] (label_ackley200) {\textbf{Ackley 200D}};
                \node[anchor=west, yshift=-9.5cm, xshift=0.08cm] (ackley200) {
                    \includegraphics[scale=0.90]{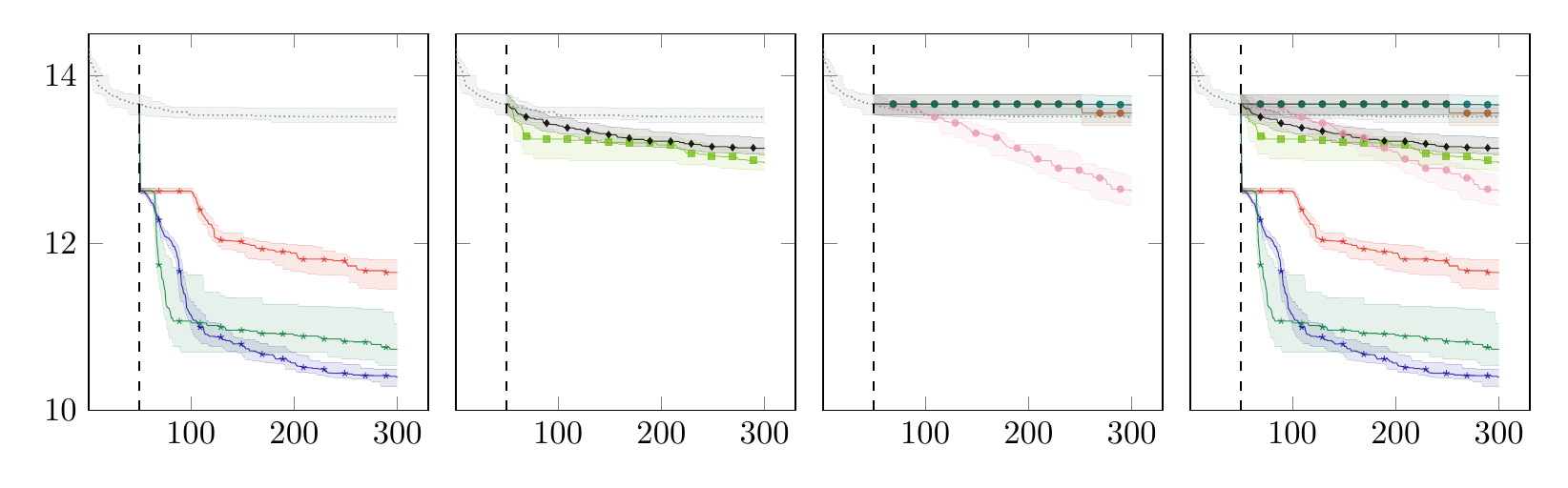}
                };

                \node[anchor=north west, yshift=-11.7cm, xshift=0.5cm] (label_robot14) {\textbf{Robot 14D}};
                \node[anchor=west, yshift=-14.3cm, xshift=-0.01cm] (robot14) {
                    \includegraphics[scale=0.90]{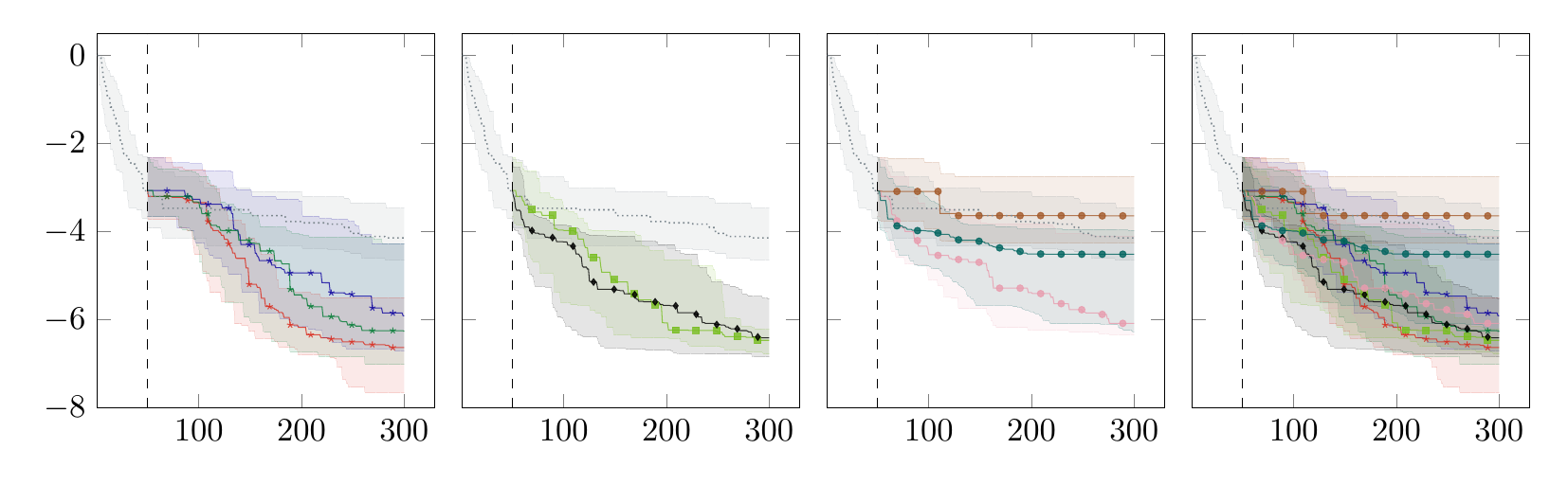}
                };

                \node[anchor=west, yshift=-17.5cm, xshift=1.2cm] (legend) {
                    \includegraphics[trim={0 0 0 5.5cm},clip,width=0.62\paperwidth]{figures/legend.pdf}
                };

                % add axis labels
                \node[anchor=south, rotate=90, yshift=-0.1cm] at (rover60.west) (labelY) {\footnotesize Objective};

                \node[anchor=north, xshift=0.4cm, yshift=0.5cm] at (rover60.south) (labelX) {\footnotesize Number of evaluations};

                \node[anchor=south, rotate=90, yshift=-0.5cm] at (ferm80.west) (labelY) {\footnotesize Objective};

                \node[anchor=north, xshift=0.4cm, yshift=0.5cm] at (ferm80.south) (labelX) {\footnotesize Number of evaluations};

                \node[anchor=south, rotate=90, yshift=-0.1cm] at (ackley200.west) (labelY) {\footnotesize Objective};

                \node[anchor=north, xshift=0.4cm, yshift=0.5cm] at (ackley200.south) (labelX) {\footnotesize Number of evaluations};

                \node[anchor=south, rotate=90, yshift=-0.2cm] at (robot14.west) (labelY) {\footnotesize Objective};

                \node[anchor=north, xshift=0.4cm, yshift=0.5cm] at (robot14.south) (labelX) {\footnotesize Number of evaluations};
            \end{tikzpicture}
        \end{center}
        \caption{Comparison of \texttt{ENTMOOT} to other black-box optimization frameworks (see:
        Section~\ref{sec:numerical_studies}) using various benchmark functions. Compared to Fig
        .~\ref{fig:entmoot_performance_study} confidence intervals are shown as well.}
        \label{fig:entmoot_perf_study_conf_app}
    \end{figure*}

    \subsection{Large-scale extension}
    Here, we give more details on the large-scale numerical studies performed in Section~\ref{sec:large_study}. A
    publicly available dataset \citep{yeh1998ModelingNetworksc} was used which describes the compressive strength of
    concrete, based on its composition and production procedure defined by features $\boldsymbol{x} \in
    \mathbb{R}^{8}$. The dataset is available at the UCI machine learning repository \citep{dua2017UCI}. We train a
    large GBT model using 4000 tree estimators, maximum interaction depth of 14, minimum number of samples in leaf of
    2, maximum number of leaves of 64 and random state 101. All other hyperparameter settings are left at default
    values. We do not claim that these hyperparameters make sense for the particular dataset, the point is to show
    how tailored branch-and-bound algorithms can be used to make \texttt{ENTMOOT} more scalable. \par

    We use 500 clusters for the k-means algorithm, determining the clusters for the cluster distance penalty
    introduced in Section~\ref{sec:handling_large_data}. Strong branching and partition refinement according to
    \citet{mistry2018MixedIntegerEmbedded} is used to handle the large-scale optimization problem. For the strong
    branching strategy we use a lookahead value of 200, i.e.\ the algorithm evaluates Equ.~\eqref{eq:lowerbound}
    $b^{\alpha,S}$ for the next 200 branches that would be explored and removes them if they lead to lower objective
    bounds above the best feasible solution found so far. The partition refinement is governed by the initial group
    size of 20 and a fixed bounding time of 120 s per iteration in ENTMOOT and derives Equ.~\eqref{eq:lowerbound}
    $b^{\hat{\mu},S}$. The underlying mixed-integer linear problem solved during partition refinement is handled by
    Gurobi~9. These hyperparameter specifications were not optimized and are in line with
    \citet{mistry2018MixedIntegerEmbedded}, where further details regarding the here mentioned algorithmic concepts
    can be found. Both the modified \texttt{ENTMOOT} algorithm and Gurobi~9 were tested for hyperparameters $\kappa
    \in \left\{ 0.01,0.1,1,10,100,1000 \right\}$, with a fixed time limit of 4~h. Table~\ref{tab:concres} summarizes
    the results for all penalty parameter values. Blank entries in Table~\ref{tab:concres} refer to an early
    convergence to an $\epsilon$-global optimal solution, given a relative optimality gap of 0.01 \%. For the
    modeling of the mixed-integer programs used in the modified \texttt{ENTMOOT} algorithm we used Pyomo~5.6.7
    \citep{hart2011pyomo}.

    \subsection{Runtime comparison}
    Table~\ref{tab:runtime} summarizes run times for all algorithm applied to different benchmark problems. The
    abbreviations \textit{ros}, \textit{ras}, \textit{sph} and \textit{stang} refer to Rosenbrock, Rastrigin, Sphere
    and Styblinski-Tang benchmark functions, respectively. \textit{ferm}, \textit{rover}, \textit{ackl} and
    \textit{robot} denote fermentation, rover trajectory planning, Ackley and robot push benchmark problems,
    respectively. The number in benchmark function names indicates the dimensionality of the problem, e.g.\
    \textit{ackl200} refers to Ackley 200D. Algorithms names \textit{E-GLOB}, \textit{E-RND}, \textit{E-L1} and
    \textit{E-LARGE} denote \texttt{ENTMOOT}, \texttt{ENTMOOT-RND}, \texttt{ENTMOOT-L1} and \texttt{ENTMOOT-LARGE},
    respectively. \texttt{SKOPT-GBRT}, \texttt{SKOPT-RF} and \texttt{SKOPT-GP} are abbreviated by \textit{S-GBRT},
    \textit{S-RF} and \textit{S-GP}. \texttt{BOHAMIANN} is denoted by \textit{BOH}. \par
    Numerical entries in Table~\ref{tab:runtime} indicate $mean(\text{runtime}) \pm standard deviation
    (\text{runtime})$ derived from all runs presented in the paper. Units are seconds (s), minutes (m) and hours (h).
    The benchmark studies presented here focus on comparing the quality of solutions derived by individual algorithm.
    The assumption is that evaluating the black-box function is way more expensive than running an iteration the
    black-box optimization algorithm. Therefore, \texttt{ENTMOOT} optimized to find good solutions for benchmark
    problems but not to minimize its run time. However, when comparing run times \texttt{ENTMOOT} performs similar to
    other sophisticated algorithms like \texttt{BOHAMIANN} and \texttt{SKOPT-GP}. Also, it is important to emphasize
    that \texttt{ENTMOOT}'s run time can be reduced by lowering the time limit of Gurobi. This may lead to worse
    black-box function values as the tree-based model may not be optimized to global optimality.

    \setlength\tabcolsep{2.3pt}
    \renewcommand{\arraystretch}{1.5}
    \begin{table}[]
        \scriptsize
        \begin{center}
            \begin{tabular}{l||c|c|c|c|c|c|c||c|c|c|c|c||c}
                & E-GLOB & E-RND & E-L1 & E-LARGE & SMAC & S-GBRT & S-RF & BOH & S-GP & CMA & BFGS & NM & DUMMY \\
                \hline
                \hline
                ros10 & 5m $\pm$ 0 & 1m $\pm$ 0 & 1.8h $\pm$ 0.2 & 5m $\pm$ 1 & 5m $\pm$ 0 & 38s $\pm$ 1 & 1m $\pm$ 0
                & -
                & -
                & -
                & -
                & -
                & 4s $\pm$ 2
                \\
                ros20 & 6m $\pm$ 1 & 1m $\pm$ 0 & 4.9h $\pm$ 0.2 & 8m $\pm$ 1 & 8m $\pm$ 0 & 50s $\pm$ 2 & 1m $\pm$ 0
                & -
                & -
                & -
                & -
                & -
                & 4s $\pm$ 1
                \\
                ros40 & 26m $\pm$ 10 & 2m $\pm$ 0 & 4.2h $\pm$ 0.5 & 0.6h $\pm$ 0.1 & 16m $\pm$ 1 & 1m $\pm$ 0
                & 2m $\pm$ 0
                & -
                & -
                & -
                & -
                & -
                & 4s $\pm$ 1
                \\
                ras10 & 8m $\pm$ 1 & 1m $\pm$ 0 & 4.8h $\pm$ 0.3 & 8m $\pm$ 1 & 4m $\pm$ 0 & 38s $\pm$ 1 & 59s $\pm$ 1
                & -
                & -
                & -
                & -
                & -
                & 4s $\pm$ 2
                \\
                ras20 & 2.5h $\pm$ 0.8 & 1m $\pm$ 0 & 7.8h $\pm$ 0.1 & 1.4h $\pm$ 0.5 & 8m $\pm$ 0 & 1m $\pm$ 1
                & 1m $\pm$ 0
                & -
                & -
                & -
                & -
                & -
                & 4s $\pm$ 1
                \\
                ras40 & 6.4h $\pm$ 0.1 & 2m $\pm$ 0 & 7.7h $\pm$ 0.1 & 6.7h $\pm$ 0.1 & 16m $\pm$ 2 & 1m $\pm$ 0
                & 2m $\pm$ 0
                & -
                & -
                & -
                & -
                & -
                & 5s $\pm$ 1
                \\
                sph10 & 0.8h $\pm$ 0.2 & 1m $\pm$ 0 & 6.2h $\pm$ 0.2 & 0.6h $\pm$ 0.2 & 5m $\pm$ 0 & 38s $\pm$ 1
                & 1m $\pm$ 0
                & -
                & -
                & -
                & -
                & -
                & 4s $\pm$ 2
                \\
                sph20 & 5.2h $\pm$ 0.2 & 1m $\pm$ 0 & 7.6h $\pm$ 0.1 & 5.0h $\pm$ 0.2 & 8m $\pm$ 0 & 50s $\pm$ 2
                & 1m $\pm$ 0
                & -
                & -
                & -
                & -
                & -
                & 3s $\pm$ 1
                \\
                sph40 & 5.9h $\pm$ 0.2 & 2m $\pm$ 0 & 7.7h $\pm$ 0.1 & 6.1h $\pm$ 0.1 & 16m $\pm$ 1 & 1m $\pm$ 0
                & 2m $\pm$ 0
                & -
                & -
                & -
                & -
                & -
                & 4s $\pm$ 1
                \\
                stang10 & 8m $\pm$ 1 & 1m $\pm$ 0 & 2.9h $\pm$ 0.3 & 10m $\pm$ 1 & 5m $\pm$ 0 & 37s $\pm$ 1 & 1m $\pm$ 0
                & -
                & -
                & -
                & -
                & -
                & 3s $\pm$ 1
                \\
                stang20 & 12m $\pm$ 2 & 1m $\pm$ 0 & 6.0h $\pm$ 0.1 & 15m $\pm$ 2 & 8m $\pm$ 0 & 48s $\pm$ 1
                & 1m $\pm$ 0
                & -
                & -
                & -
                & -
                & -
                & 4s $\pm$ 2
                \\
                stang40 & 1.0h $\pm$ 0.3 & 2m $\pm$ 0 & 5.7h $\pm$ 0.2 & 1.7h $\pm$ 0.4 & 17m $\pm$ 1 & 1m $\pm$ 0
                & 2m $\pm$ 1
                & -
                & -
                & -
                & -
                & -
                & 5s $\pm$ 2
                \\
                \hline
                ferm80 & 4.4h $\pm$ 1.2 & 3m $\pm$ 0 & 8.4h $\pm$ 0.0 & 1.5h $\pm$ 1.5 & 0.6h $\pm$ 0.0 & 2m $\pm$ 0
                & 3m $\pm$ 0
                & 5.8h $\pm$ 0.1
                & 1.3h $\pm$ 0.2
                & 14s $\pm$ 1
                & 17s $\pm$ 2
                & 15s $\pm$ 2
                & 14s $\pm$ 1
                \\
                rover60 & 5m $\pm$ 1 & 2m $\pm$ 1 & 9m $\pm$ 1 & 6m $\pm$ 1 & 25m $\pm$ 1 & 2m $\pm$ 0 & 2m $\pm$ 0
                & 5.7h $\pm$ 0.1
                & 1.3h $\pm$ 0.1
                & 5s $\pm$ 2
                & 5s $\pm$ 2
                & 4s $\pm$ 1
                & 4s $\pm$ 1
                \\
                ackl200 & 7m $\pm$ 2 & 5m $\pm$ 0 & 11m $\pm$ 1 & 7m $\pm$ 2 & 2.1h $\pm$ 0.1 & 5m $\pm$ 0 & 6m $\pm$ 0
                & 6.3h $\pm$ 0.1
                & 8.0h $\pm$ 0.8
                & 4s $\pm$ 2
                & 4s $\pm$ 1
                & 4s $\pm$ 1
                & 7s $\pm$ 2
                \\
                robot14 & 23m $\pm$ 12 & 1m $\pm$ 0 & 1.3h $\pm$ 1.2 & 18m $\pm$ 9 & 6m $\pm$ 0 & 43s $\pm$ 2
                & 1m $\pm$ 0
                & 5.5h $\pm$ 0.1
                & 29m $\pm$ 3
                & 5s $\pm$ 2
                & 5s $\pm$ 1
                & 4s $\pm$ 1
                & 4s $\pm$ 1
                \\
            \end{tabular}
        \end{center}
        \caption{\label{tab:runtime}Shows all run times of different tools used for benchmarking. Due to space
        restrictions we abbreviate both algorithms and benchmark problems.}
    \end{table}

    \subsection{Computational setup}
    All numerical results are computed using a Linux machine with 16 GB RAM and an Intel Core i7-7700K @ 4.20 GHz CPU
    in a \texttt{HTCondor} \citep{thain2005Condor} setup to allow running multiple experiments in parallel.

%\clearpage

\bibliography{citations}
\end{document}